\documentclass[10pt,journal,compsoc]{IEEEtran}
\ifCLASSOPTIONcompsoc
 \usepackage{cite}
\else
 \usepackage{cite}
\fi

\ifCLASSINFOpdf
\else
\fi
\usepackage{amsmath,amssymb}
\usepackage{mathrsfs} 
\usepackage[dvipsnames]{xcolor}
\usepackage{multirow}
\usepackage{graphicx}
\newcommand{\cut}[1]{}
\newcommand{\keypoint}[1]{\vspace{0.1cm}\noindent\textbf{#1}\quad}

\usepackage{color, colortbl}
\definecolor{Gray}{gray}{0.85}
\DeclareMathOperator*{\E}{\mathbb{E}}

\newcommand{\red}[1]{\textcolor{red}{#1}} 
\newcommand{\blue}[1]{\textcolor{blue}{#1}}
\newcommand{\green}[1]{\textcolor{ForestGreen}{#1}}
\newcommand{\purple}[1]{\textcolor{Plum}{#1}}

\newcommand{\argmin}[1]{\underset{#1}{\operatorname{arg}\,\operatorname{min}}\;}
\newcommand{\etal}{\textit{et al}.}

\usepackage{verbatim}

\begin{document}

\title{Meta-Learning in Neural Networks: A Survey}

\author{Timothy Hospedales, Antreas Antoniou, Paul Micaelli, Amos Storkey\thanks{T. Hospedales is with Samsung AI Centre, Cambridge and University of Edinburgh. A. Antoniou, P. Micaelli and Storkey are with University of Edinburgh. Email: \{t.hospedales,a.antoniou,paul.micaelli,a.storkey\}@ed.ac.uk.}}

\IEEEtitleabstractindextext{%
\begin{abstract}

The field of meta-learning, or learning-to-learn, has seen a dramatic rise in interest in recent years. Contrary to conventional approaches to AI where tasks are solved from scratch using a fixed learning algorithm, meta-learning aims to improve the learning algorithm itself, given the experience of multiple learning episodes. This paradigm provides an opportunity to tackle many conventional challenges of deep learning, including data and computation bottlenecks, as well as generalization. This survey describes the contemporary meta-learning landscape. We first discuss definitions of meta-learning and position it with respect to related fields, such as transfer learning and hyperparameter optimization. We then propose a new taxonomy that provides a more comprehensive breakdown of the space of meta-learning methods today. We survey promising applications and successes of meta-learning such as few-shot learning and reinforcement learning. Finally, we discuss outstanding challenges and promising areas for future research.

\end{abstract}

}

\maketitle
\IEEEdisplaynontitleabstractindextext
\IEEEpeerreviewmaketitle

\section{Introduction}\label{sec:introduction}

Contemporary machine learning models are typically trained from scratch for a specific task using a fixed learning algorithm designed by hand. Deep learning-based approaches specifically have seen great successes in a variety of fields \cite{he2016resnet,silver2016mastering,devlin2019bert}. However there are clear limitations \cite{marcus2018deepcritical}. For example, successes have largely been in areas where vast quantities of data can be collected or simulated, and where huge compute resources are available. This excludes many applications where data is intrinsically rare or expensive \cite{altae2016oneshotdrugdiscovery}, or compute resources are unavailable \cite{ignatov2019aismartphone}.

Meta-learning provides an alternative paradigm where a machine learning model gains experience over multiple learning episodes -- often covering a distribution of related tasks -- and uses this experience to improve its future learning performance. 
This `learning-to-learn' \cite{thrun1998learning} can lead to a variety of benefits such as data and compute efficiency, and it is better aligned with human and animal learning \cite{harlow1949formation}, where learning strategies improve both on a lifetime and evolutionary timescales \cite{biggs1985metarole,harlow1949formation,schrier1984learning}.

Historically, the success of machine learning was driven by the choice of hand-engineered features \cite{domingos2012usefulml, lowe2004sift}. Deep learning realised the promise of joint feature and model learning \cite{krizhevsky2012imagenetdeepcnn}, providing a huge improvement in performance for many tasks \cite{he2016resnet,devlin2019bert}. Meta-learning in neural networks can be seen as aiming to provide the next step of integrating joint feature, model, and \emph{algorithm} learning.

Neural network meta-learning has a long history \cite{schmidhuber1987evolutionary,schmidhuber1997inductivebias,thrun1998learning}. However, its potential as a driver to advance the frontier of the contemporary deep learning industry has led to an explosion of recent research. In particular meta-learning has the potential to alleviate many of the main criticisms of contemporary deep learning \cite{marcus2018deepcritical}, for instance by improving data efficiency, knowledge transfer and unsupervised learning. Meta-learning has proven useful both in multi-task scenarios where task-agnostic knowledge is extracted from a family of tasks and used to improve learning of new tasks from that family \cite{finn2017maml,thrun1998learning}; and single-task scenarios where a single problem is solved repeatedly and improved over multiple \emph{episodes} \cite{franceschi2018bilevel,liu2019darts,andrychowicz2016graddescgraddesc}.
Successful applications have been demonstrated in areas spanning few-shot image recognition \cite{finn2017maml,snell2017prototypicalnets}, unsupervised learning \cite{metz2018meta}, data efficient \cite{duan2016fastslowrl,houthooft2018epg} and self-directed \cite{alet2020metacuriosity} reinforcement learning (RL), hyperparameter optimization \cite{franceschi2018bilevel}, and neural architecture search (NAS) \cite{liu2019darts, real2018regularized, zoph2016neural}.

Many perspectives on meta-learning can be found in the literature, in part because different communities use the term differently. 
{Thrun \cite{thrun1998learning} operationally defines learning-to-learn as occurring when a learner's performance at solving tasks drawn from a given task family improves with respect to the number of tasks seen. (\emph{cf.}, conventional machine learning performance improves as more data from a single task is seen).} This perspective \cite{vilalta2002perspective,thrun1998lifelong,baxter1998theoretical} views meta-learning as a tool to manage the `no free lunch' theorem \cite{wolpert1996nofreelunchlearning} and improve generalization by searching for the algorithm (inductive bias) that is best suited to a given problem, or problem family. However, this definition can include transfer, multi-task, feature-selection, and model-ensemble learning, which are not typically considered as meta-learning today. 
Another usage of meta-learning \cite{vanschoren2018metasurvey} deals with algorithm selection based on dataset features, and becomes hard to distinguish from automated machine learning (AutoML) \cite{yao2018automl,hutter2019autoMLbook}. 

In this paper, we focus on contemporary \emph{neural-network} meta-learning. We take this to mean algorithm learning as per \cite{thrun1998lifelong,vilalta2002perspective}, but focus specifically on where this is achieved by \emph{end-to-end} learning of an \emph{explicitly defined objective function} (such as cross-entropy loss). Additionally we consider single-task meta-learning, and discuss a wider variety of (meta) objectives such as robustness and compute efficiency.

This paper thus provides a unique, timely, and up-to-date survey of the rapidly growing area of neural network meta-learning. In contrast, previous surveys are rather out of date and/or focus on algorithm selection for data mining \cite{vilalta2002perspective, pan2009transfer_survey, lemke2015metalearning,  vanschoren2018metasurvey}, AutoML \cite{yao2018automl,hutter2019autoMLbook}, or particular applications of meta-learning such as few-shot learning \cite{wang2019fewshotsurvey} or neural architecture search \cite{elsken2019nas}.

We address both meta-learning methods and applications. We first introduce meta-learning through a high-level problem formalization that can be used to understand and position work in this area. We then provide a new taxonomy in terms of meta-representation, meta-objective and meta-optimizer. This framework provides a design-space for developing new meta learning methods and customizing them for different applications. We survey several popular and emerging application areas including few-shot, reinforcement learning, and architecture search; and position meta-learning with respect to related topics such as transfer and multi-task learning. We conclude by discussing outstanding challenges and areas for future research.

\cut{\input{AlternativeFormalism}}

\section{Background}


Meta-learning is difficult to define, having been used in various inconsistent ways, even within contemporary neural-network literature. In this section, we introduce our definition and key terminology, and then position meta-learning with respect to related topics. 

Meta-learning is most commonly understood as \emph{learning to learn}, which refers to the process of improving a learning algorithm over multiple learning episodes. In contrast, conventional ML improves model predictions over multiple data instances. During \textbf{base learning}, an \emph{inner} (or \emph{lower}/\emph{base}) learning algorithm solves a \emph{task} such as image classification \cite{krizhevsky2012imagenetdeepcnn}, defined by a dataset and objective. During \textbf{meta-learning}, an \emph{outer} (or \emph{upper}/\emph{meta}) algorithm updates the inner learning algorithm such that the model it learns improves an outer objective. For instance this objective could be generalization performance or learning speed of the inner algorithm. Learning episodes of the base task, namely (base algorithm, trained model, performance) tuples, can be seen as providing the instances needed by the outer algorithm to learn the base learning algorithm.

As defined above, many conventional algorithms such as random search of hyper-parameters by cross-validation could fall within the definition of meta-learning. The salient characteristic of contemporary neural-network meta-learning is an explicitly defined \emph{meta-level objective}, and \emph{end-to-end} optimization of the inner algorithm with respect to this objective. Often, meta-learning is conducted on learning episodes sampled from a task family, leading to a base learning algorithm that performs well on new tasks sampled from this family. However, in a limiting case all training episodes can be sampled from a single task. In the following section, we introduce these notions more formally.

\subsection{Formalizing Meta-Learning}\label{sec:formalize}

\keypoint{Conventional Machine Learning}
In conventional supervised machine learning, we are given a training dataset $\mathcal{D}=\{(x_1,y_1),\dots,(x_N,y_N)\}$, such as (input image, output label) pairs. We can train a predictive model $\hat{y}=f_\theta(x)$ parameterized by $\theta$, by solving:
\begin{align}
\begin{split}
\label{eq:supervised}
\theta^* &= \arg\min_{\theta} \mathcal{L}(\mathcal{D};\theta,\omega)
\end{split}
\end{align}

\noindent where $\mathcal{L}$ is a loss function that measures the error between true labels and those predicted by $f_\theta(\cdot)$.
The conditioning on $\omega$ denotes the dependence of this solution on assumptions about `how to learn', such as the choice of optimizer for $\theta$ or function class for $f$. Generalization is then measured by evaluating a number of test points with known labels.

The conventional assumption is that this optimization is performed \emph{from scratch} for every problem $\mathcal{D}$; and that $\omega$ is pre-specified. However, the specification of $\omega$ can drastically affect performance measures like accuracy or data efficiency. Meta-learning seeks to improve these measures by learning the learning algorithm itself, rather than assuming it is pre-specified and fixed. This is often achieved by revisiting the first assumption above, and learning from a distribution of tasks rather than from scratch.

\keypoint{Meta-Learning: Task-Distribution View} A common view of meta-learning is to learn a general purpose learning algorithm that can generalize across tasks, and ideally enable each new task to be learned better than the last. We can evaluate the performance of $\omega$ over a distribution of tasks $p(\mathcal{T})$. Here we loosely define a task to be a dataset and loss function $\mathcal{T} = \{\mathcal{D}, \mathcal{L} \}$. Learning how to learn thus becomes

\begin{align}
\label{eq:metaAbstract}
 \min_\omega \E_{\mathcal{T}\sim p(\mathcal{T})} \mathcal{L(\mathcal{D};\omega)}
\end{align}

\noindent where $\mathcal{L}(\mathcal{D};\omega)$ measures the performance of a model trained using $\omega$ on dataset $\mathcal{D}$. `How to learn', i.e. $\omega$, is often referred to as \emph{across-task} knowledge or \emph{meta-knowledge}.

To solve this problem in practice, we often assume access to a \emph{set} of source tasks sampled from $p(\mathcal{T})$. Formally, we denote the set of $M$ source tasks used in the meta-training stage as $\mathscr{D}_{source}=\{(\mathcal{D}^{train}_{source}, \mathcal{D}^{val}_{source})^{(i)}\}_{i=1}^M$ where each task has both training and validation data. Often, the source train and validation datasets are respectively called \emph{support} and \emph{query} sets. The \textbf{meta-training} step of `learning how to learn' can be written as:

\begin{align}
\label{eq:metatrain}
 \omega^* &=\arg\max_\omega\log p(\omega|\mathscr{D}_{source})
\end{align}

Now we denote the set of $Q$ target tasks used in the meta-testing stage as $\mathscr{D}_{target}=\{(\mathcal{D}^{train}_{target}, \mathcal{D}^{test}_{target})^{(i)}\}_{i=1}^Q$ where each task has both training and test data. In the \textbf{meta-testing} stage\cut{\footnote{While we use meta-train and meta-test to refer to the phases outlined above, we note that some papers overload these terms to refer to the inner and outer optimizations respectively within meta-training as used above.}} we use the learned meta-knowledge $\omega^*$ to train the base model on each previously unseen target task $i$:

\begin{equation}\label{eq:metatest}
 \theta^{*~(i)} = \arg\max_\theta\log p(\theta|\omega^*,\mathcal{D}^{train~(i)}_{target})
\end{equation}

In contrast to conventional learning in Eq.~\ref{eq:supervised}, learning on the training set of a target task $i$ now benefits from meta-knowledge $\omega^*$ about the algorithm to use. This could be an estimate of the initial parameters \cite{finn2017maml}, or an entire learning model \cite{mishra2017simple} or optimization strategy \cite{ravi2016optimization}. We can evaluate the accuracy of our meta-learner by the performance of $\theta^{*~(i)}$ on the test split of each target task $\mathcal{D}^{test~(i)}_{target}$.

This setup leads to analogies of conventional underfitting and overfitting: \emph{meta-underfitting} and \emph{meta-overfitting}. In particular, meta-overfitting is an issue whereby the meta-knowledge learned on the source tasks does not generalize to the target tasks. It is relatively common, especially in the case where only a small number of source tasks are available. It can be seen as learning an inductive bias $\omega$ that constrains the hypothesis space of $\theta$ too tightly around solutions to the source tasks.

\keypoint{Meta-Learning: Bilevel Optimization View} The previous discussion outlines the common flow of meta-learning in a multiple task scenario, but does not specify how to solve the meta-training step in Eq.~\ref{eq:metatrain}. This is commonly done by casting the meta-training step as a bilevel optimization problem. While this picture is arguably only accurate for the optimizer-based methods (see section \ref{sec:alg}), it is helpful to visualize the mechanics of meta-learning more generally. Bilevel optimization \cite{stackelberg1952bileveloptimization} refers to a hierarchical optimization problem, where one optimization contains another optimization as a constraint \cite{franceschi2018bilevel,sinha2018bilevelsurvey}. Using this notation, meta-training can be formalised as follows:

\begin{align}
 \omega^* &= \argmin{\omega} \sum^M_{i=1}\mathcal{L}^{meta}(\theta^{*~(i)}(\omega), \omega, \mathcal{D}^{val~(i)}_{source})\label{eq:upper}\\
 \text{s.t. }\theta^{*(i)}(\omega) &= \argmin{{\theta}}\mathcal{L}^{task}(\theta, \omega, \mathcal{D}^{train~(i)}_{source})\label{eq:lower}
\end{align}

\noindent where $\mathcal{L}^{meta}$ and $\mathcal{L}^{task}$ refer to the outer and inner objectives respectively, such as cross entropy in the case of few-shot classification. Note the leader-follower asymmetry between the outer and inner levels: the inner level optimization Eq.~\ref{eq:lower} is conditional on the learning strategy $\omega$ defined by the outer level, but it cannot change $\omega$ during its training.

Here $\omega$ could indicate an initial condition in non-convex optimization \cite{finn2017maml}, a hyper-parameter such as regularization strength \cite{franceschi2018bilevel}, or even a parameterization of the loss function to optimize $\mathcal{L}^{task}$ \cite{li2019featurecritic}. Section~\ref{sec:what} discusses the space of choices for $\omega$ in detail. The outer level optimization learns $\omega$ such that it produces models $\theta^{*~(i)}(\omega)$ that perform well on their validation sets after training. Section~\ref{sec:metaOpt} discusses how to optimize $\omega$ in detail. In Section~\ref{sec:why} we consider what $\mathcal{L}^{meta}$ can measure, such as validation performance, learning speed or model robustness.

Finally, we note that the above formalization of meta-training uses the notion of a distribution over tasks. While common in the meta-learning literature, it is not a necessary condition for meta-learning. More formally, if we are given a single train and test dataset ($M=Q=1$), we can split the training set to get validation data such that $\mathscr{D}_{source} = (\mathcal{D}^{train}_{source}, \mathcal{D}^{val}_{source})$ for meta-training, and for meta-testing we can use $\mathscr{D}_{target} = (\mathcal{D}^{train}_{source} \cup \mathcal{D}^{val}_{source}, \mathcal{D}^{test}_{target})$. We still learn $\omega$ over several episodes, and different train-val splits are usually used during meta-training.

\keypoint{Meta-Learning: Feed-Forward Model View} As we will see, there are a number of meta-learning approaches that synthesize models in a feed-forward manner, rather than via an explicit iterative optimization as in Eqs.~\ref{eq:upper}-\ref{eq:lower} above. While they vary in their degree of complexity, it can be instructive to understand this family of approaches by instantiating the abstract objective in Eq.~\ref{eq:metaAbstract} to define a toy example for meta-training linear regression \cite{denevi2018l2lmean}. 

\begin{align}
\label{eq:metaBBM}
 \min_\omega \E_{\substack{
\mathcal{T}\sim p(\mathcal{T})\\
(\mathcal{D}^{tr},\mathcal{D}^{val})\in \mathcal{T}}}~~\sum_{(\mathbf{x},y)\in\mathcal{D}^{val}}\left[ (\mathbf{x}^T \mathbf{g}_\omega(\mathcal{D}^{tr})-y)^2\right]
\end{align}

\noindent Here we meta-train by optimizing over a distribution of tasks. For each task a train and validation set is drawn. The train set $\mathcal{D}^{tr}$ is embedded \cite{zaheer2017deepSet} into a vector $\mathbf{g}_\omega$ which defines the linear regression weights to predict examples $\mathbf{x}$ from the validation set. Optimizing Eq.~\ref{eq:metaBBM} `learns  to learn' by training the function $\mathbf{g}_\omega$ to map a training set to a weight vector. Thus $\mathbf{g}_\omega$ should provide a good solution for novel meta-test tasks $\mathcal{T}^{te}$ drawn from $p(\mathcal{T})$. Methods in this family vary in the complexity of the predictive model $\mathbf{g}$ used, and how the support set is embedded \cite{zaheer2017deepSet} (e.g., by pooling, CNN or RNN). These models are also known as \emph{amortized} \cite{gordon2018metaprobainference} because the cost of learning a new task is reduced to a feed-forward operation through $\mathbf{g}_\omega(\cdot)$, with iterative optimization already paid for during meta-training of $\omega$.

\subsection{Historical Context of Meta-Learning}
Meta-learning and learning-to-learn first appear in the literature in 1987\cite{schmidhuber1987evolutionary}. J. Schmidhuber introduced a family of methods that can learn how to learn, using \emph{self-referential} learning. Self-referential learning involves training neural networks that can receive as inputs their own weights and predict updates for said weights. Schmidhuber proposed to learn the model itself using evolutionary algorithms.

Meta-learning was subsequently extended to multiple areas. Bengio \etal \cite{bengio1990learning,bengio1995} proposed to meta-learn biologically plausible learning rules. Schmidhuber \etal continued to explore self-referential systems and meta-learning \cite{schmidhuber1996simple,schmidhuber1993neural}. S. Thrun \etal~took care to more clearly define the term \emph{learning to learn} in \cite{thrun1998learning} and introduced initial theoretical justifications and practical implementations. 
Proposals for training meta-learning systems using gradient descent and backpropagation were first made in 1991 \cite{schmidhuber1991possibility} followed by more extensions in 2001 \cite{hochreiter2001learning,younger2001meta}, with \cite{vilalta2002perspective} giving an overview of the literature at that time. Meta-learning was used in the context of reinforcement learning in 1995 \cite{storck1995reinforcement}, followed by various extensions \cite{wiering1998efficient, schweighofer2003meta}.

\subsection{Related Fields}
Here we position meta-learning against related areas whose relation to meta-learning is often a source of confusion.

\keypoint{Transfer Learning (TL)} TL \cite{pan2009transfer_survey,pratt1991direct} 
uses past experience from a source task to improve learning (speed, data efficiency, accuracy) on a target task. TL refers both to this problem area and family of solutions, most commonly parameter transfer plus optional fine tuning \cite{yosinski2014howtransferable} (although there are numerous other approaches \cite{pan2009transfer_survey}).

In contrast, meta-learning refers to a paradigm that can be used to improve TL as well as other problems. In TL the prior is extracted by vanilla learning on the source task without the use of a meta-objective. In meta-learning, the corresponding prior would be defined by an outer optimization that evaluates the benefit of the prior when learn a new task, as illustrated by MAML \cite{finn2017maml}. More generally, meta-learning deals with a much wider range of meta-representations than solely model parameters (Section~\ref{sec:what}).

\keypoint{Domain Adaptation (DA) and Domain Generalization (DG)}
Domain-shift refers to the situation where source and target problems share the same objective, but the input distribution of the target task is shifted with respect to the source task \cite{csurka2017domainadaptationbook,pan2009transfer_survey}, reducing model performance. DA is a variant of transfer learning that attempts to alleviate this issue by adapting the source-trained model using sparse or unlabeled data from the target. DG refers to methods to train a source model to be robust to such domain-shift without further adaptation. Many knowledge transfer methods have been studied \cite{csurka2017domainadaptationbook,pan2009transfer_survey} to boost target domain performance. However, as for TL, vanilla DA and DG don't use a meta-objective to optimize `how to learn' across domains. Meanwhile, meta-learning methods can be used to perform both DA \cite{li2020metadomain} and DG \cite{li2019featurecritic} (see Sec.~\ref{subsec:domainshift}).

\keypoint{Continual learning (CL)} Continual or lifelong learning \cite{ring1994continuallearningreinforcement, parisi2019continualsurvey,chen2018lifelong} refers to the ability to learn on a sequence of tasks drawn from a potentially non-stationary distribution, and in particular seek to do so while accelerating learning new tasks and without forgetting old tasks. Similarly to meta-learning, a task distribution is considered, and the goal is partly to accelerate learning of a target task. However most continual learning methodologies are not meta-learning methodologies since this meta objective is not solved for explicitly. Nevertheless, meta-learning provides a potential framework to advance continual learning, and a few recent studies have begun to do so by developing meta-objectives that encode continual learning performance \cite{alshedivat2018nonstationarymeta,ritter2018episodicmeta,clavera2019learning}.

\keypoint{Multi-Task Learning (MTL)} aims to jointly learn several related tasks, to benefit from regularization due to parameter sharing and the diversity of the resulting shared representation \cite{rich1997multitask,yang2017deepmtrl,meyerson2019modular}, as well as compute/memory savings. Like TL, DA, and CL, conventional MTL is a single-level optimization without a meta-objective. Furthermore, the goal of MTL is to solve a fixed number of known tasks, whereas the point of meta-learning is often to solve unseen future tasks. Nonetheless, meta-learning can be brought in to benefit MTL, e.g. by learning the relatedness between tasks \cite{franceschi2017hyperopt}, or how to prioritise among multiple tasks \cite{xingyu2019auxtask}.

\keypoint{Hyperparameter Optimization (HO)} is within the remit of meta-learning, in that hyperparameters like learning rate or regularization strength describe `how to learn'. Here we include HO tasks that define a meta objective that is trained end-to-end with neural networks, such as gradient-based hyperparameter learning \cite{franceschi2017hyperopt, micaelli2020nongreedy} and neural architecture search \cite{liu2019darts}. But we exclude other approaches like random search \cite{bergstra2012randomsearchhyper} and Bayesian Hyperparameter Optimization \cite{shahriari2016bayesianoptimizationsurvey}, which are rarely considered to be meta-learning.

\keypoint{Hierarchical Bayesian Models (HBM)}
\label{sect:hbm}
involve Bayesian learning of parameters $\theta$ under a prior $p(\theta|\omega)$. The prior is written
as a conditional density on some other variable $\omega$ which has its own prior $p(\omega)$. Hierarchical Bayesian models feature strongly as models for grouped data $\mathcal{D}=\{\mathcal{D}_i|i=1,2,\ldots,M\}$, where each group $i$ has its own $\theta_i$. The full model is $\left[\prod_{i=1}^M p(\mathcal{D}_i|\theta_i)p(\theta_i|\omega)\right]p(\omega)$. The levels of hierarchy can be increased further; in particular $\omega$ can itself be parameterized, and hence $p(\omega)$ can be learnt. Learning is usually full-pipeline, but using some form of Bayesian marginalisation to compute the posterior over $\omega$: $P(\omega|\mathcal{D}) \sim p(\omega)\prod_{i=1}^M \int d\theta_i p(\mathcal{D}_i|\theta_i)p(\theta_i|\omega)$. The ease of doing the marginalisation depends on the model: in some (e.g. Latent Dirichlet Allocation \cite{blei2003lda}) the marginalisation is exact due to the choice of conjugate exponential models, in others (see e.g. \cite{edwards2016neuralstat}), a stochastic variational approach is used to calculate an approximate posterior, from which a lower bound to the marginal likelihood is computed.

Bayesian hierarchical models provide a valuable viewpoint for meta-learning, by providing a modeling rather than an algorithmic framework for understanding the meta-learning process. In practice, prior work in HBMs has typically focused on learning simple tractable models $\theta$ while most meta-learning work considers complex inner-loop learning processes, involving many iterations. Nonetheless, some meta-learning methods like MAML \cite{finn2017maml} can be understood through the lens of HBMs \cite{grant2018bayesmaml}.

\keypoint{AutoML:} AutoML \cite{yao2018automl,vanschoren2018metasurvey,hutter2019autoMLbook} is a rather broad umbrella for approaches aiming to automate parts of the machine learning process that are typically manual, such as data preparation, algorithm selection, hyper-parameter tuning, and architecture search. AutoML often makes use of numerous heuristics outside the scope of meta-learning as defined here, and focuses on tasks such as data cleaning that are less central to meta-learning. However, AutoML sometimes makes use of end-to-end optimization of a meta-objective, so meta-learning can be seen as a specialization of AutoML.

\section{Taxonomy}

\subsection{Previous Taxonomies}
\label{sec:alg}
Previous \cite{yao2020automated,lee2018mtnetmaml} categorizations of meta-learning methods have tended to produce a three-way taxonomy across optimization-based methods, model-based (or black box) methods, and metric-based (or non-parametric) methods.

\keypoint{Optimization} Optimization-based methods include those where the inner-level task (Eq.~\ref{eq:lower}) is literally solved as an optimization problem, and focuses on extracting meta-knowledge $\omega$ required to improve optimization performance. A famous example is MAML \cite{finn2017maml}, which aims to learn the initialization $\omega = \theta_0$, such that a small number of inner steps produces a classifier that performs well on validation data. This is also performed by gradient descent, differentiating through the updates of the base model. More elaborate alternatives also learn step sizes \cite{li2017meta,antoniou2018maml} or train recurrent networks to predict steps from gradients \cite{ravi2016optimization,li2017learntooptimize,andrychowicz2016graddescgraddesc}. Meta-optimization by gradient over long inner optimizations leads to several compute and memory challenges which are discussed in Section \ref{sec:challenges}. A unified view of gradient-based meta learning expressing many existing methods as special cases of a generalized inner loop meta-learning framework has been proposed \cite{grefenstette2019generalized}. 

\keypoint{Black Box / Model-based} In model-based (or black-box) methods the inner learning step (Eq.~\ref{eq:lower}, Eq.~\ref{eq:metatest}) is wrapped up in the feed-forward pass of a single model, as illustrated in Eq.~\ref{eq:metaBBM}. The model embeds the current dataset $\mathcal{D}$ into activation state, with predictions for test data being made based on this state. Typical architectures include recurrent networks \cite{hochreiter2001learning, ravi2016optimization}, convolutional networks \cite{mishra2017simple} or hypernetworks \cite{qiao2017parametersfromactivations, gidaris2018dynamicfewshot} that embed training instances and labels of a given task to define a predictor for test samples. In this case all the inner-level learning is contained in the activation states of the model and is entirely feed-forward. Outer-level learning is performed with $\omega$ containing the CNN, RNN or hypernetwork parameters. The outer and inner-level optimizations are tightly coupled as $\omega$ and $\mathcal{D}$ directly specify $\theta$. Memory-augmented neural networks \cite{graves2014neuralturing} use an explicit storage buffer and can be seen as a model-based algorithm \cite{santoro2016metamemory,munkhdalai2017metanets}. Compared to optimization-based approaches, these enjoy simpler optimization without requiring second-order gradients. However, it has been observed that model-based approaches are usually less able to generalize to out-of-distribution tasks than optimization-based methods \cite{finn2018metauniversal}. Furthermore, while they are often very good at data efficient few-shot learning, they have been criticised for being asymptotically weaker \cite{finn2018metauniversal} as they struggle to embed a large training set into a rich base model.

\keypoint{Metric-Learning} Metric-learning or non-parametric algorithms are thus far largely restricted to the popular but specific few-shot application of meta-learning (Section~\ref{sec:fsl}). The idea is to perform non-parametric `learning' at the inner (task) level by simply comparing validation points with training points and predicting the label of matching training points. In chronological order, this has been achieved with siamese \cite{kosh2015siameseoneshot}, matching  \cite{vinyals2016matching}, prototypical  \cite{snell2017prototypicalnets}, relation  \cite{sung2018relationnet}, and graph  \cite{garcia2018fewshot} neural networks. Here  outer-level learning corresponds to metric learning (finding a feature extractor $\omega$ that represents the data suitably for comparison). As before $\omega$ is learned on source tasks, and used for target tasks.

\keypoint{Discussion} The common breakdown reviewed above does not expose all facets of interest and is insufficient to understand the connections between the wide variety of meta-learning frameworks available today. For this reason, we propose a new taxonomy in the following section.

\subsection{Proposed Taxonomy}

We introduce a new breakdown along three independent axes. For each axis we provide a taxonomy that reflects the current meta-learning landscape.

\keypoint{Meta-Representation (``What?'')} The first axis is the choice of meta-knowledge $\omega$ to meta-learn. This could be anything from initial model parameters \cite{finn2017maml} to readable code in the case of program induction \cite{bello2017neuraloptimizerrl}.

\keypoint{Meta-Optimizer (``How?'')} The second axis is the choice of optimizer to use for the outer level during meta-training (see Eq.~\ref{eq:upper}). The outer-level optimizer for $\omega$ can take a variety of forms from gradient-descent \cite{finn2017maml}, to reinforcement learning \cite{bello2017neuraloptimizerrl} and evolutionary search \cite{houthooft2018epg}.

\keypoint{Meta-Objective (``Why?'')} The third axis is the \emph{goal} of meta-learning which is determined by choice of meta-objective $\mathcal{L}^{meta}$ (Eq.~\ref{eq:upper}), task distribution $p(\mathcal{T})$, and data-flow between the two levels. Together these can customize meta-learning for different purposes such as sample efficient few-shot learning \cite{finn2017maml,mishra2017simple}, fast many-shot optimization \cite{bello2017neuraloptimizerrl,wichrowska2017learnoptscalegen}, robustness to domain-shift \cite{balaji2018metareg,li2019featurecritic}, label noise \cite{li2019learning}, and adversarial attack \cite{goldblum2019adversarially}.

Together these axes provide a design-space for meta-learning methods that can orient the development of new algorithms and customization for particular applications. Note that the base model representation $\theta$ isn't included in this taxonomy, since it is determined and optimized in a way that is specific to the application at hand.

\begin{figure*}
\centering
\includegraphics[width=0.9\textwidth]{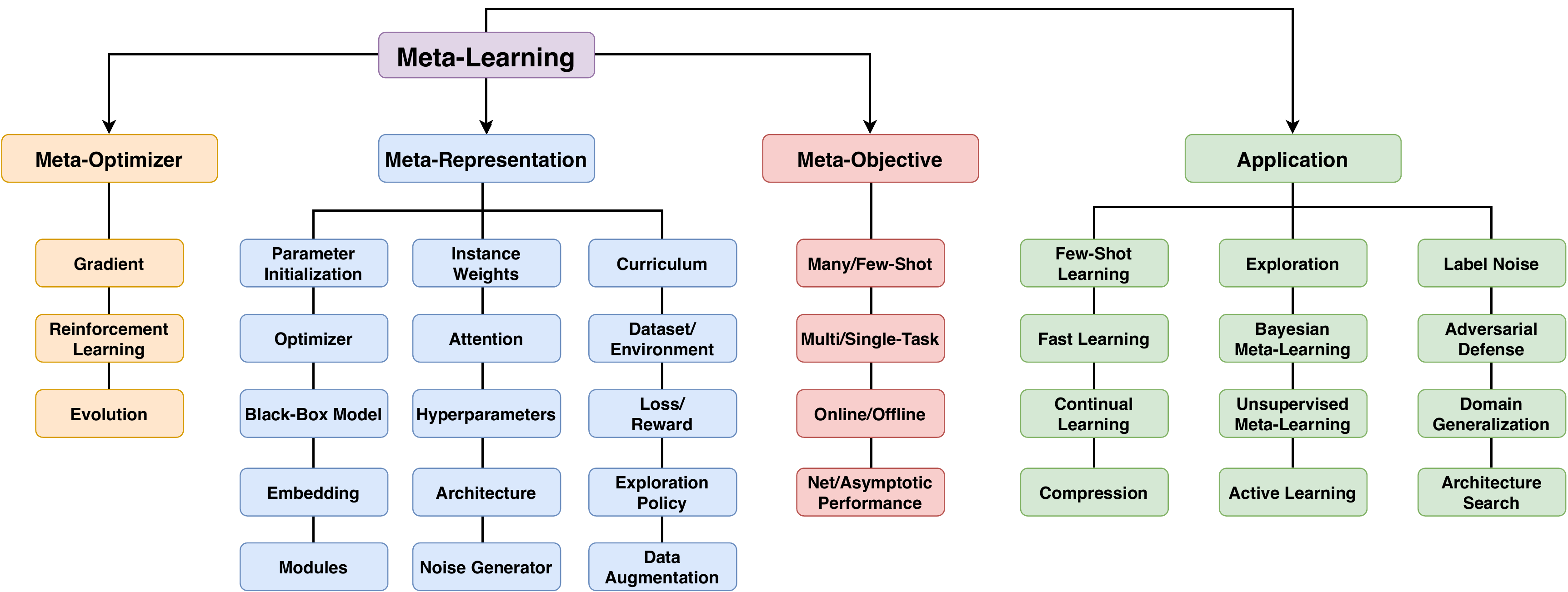}
\caption{Overview of the meta-learning landscape including algorithm design (meta-optimizer, meta-representation, meta-objective), and applications.}
\end{figure*}

\section{Survey: Methodologies}

In this section we break down existing literature according to our proposed new methodological taxonomy.

\subsection{Meta-Representation}\label{sec:what}

Meta-learning methods make different choices about what meta-knowledge $\omega$ should be, i.e. which aspects of the learning strategy should be learned; and (by exclusion) which aspects should be considered fixed.

\keypoint{Parameter Initialization} Here $\omega$ corresponds to the initial parameters of a neural network to be used in the inner optimization, with MAML being the most popular example \cite{finn2017maml,finn2018probabilistic,finn2019online}. A good initialization is just a few gradient steps away from a solution to any task $\mathcal{T}$ drawn from $p(\mathcal{T})$, and can help to learn without overfitting in few-shot learning. A key challenge with this approach is that the outer optimization needs to solve for as many parameters as the inner optimization (potentially hundreds of millions in large CNNs). This leads to a line of work on isolating a subset of parameters to meta-learn, for example by subspace \cite{lee2018mtnetmaml,rusu2019leo}, by layer \cite{antoniou2019learning,qiao2017parametersfromactivations,rusu2019leo}, or by separating out scale and shift \cite{sun2018metatransferfsl}. Another concern is whether a single initial condition is sufficient to provide fast learning for a wide range of potential tasks, or if one is limited to narrow distributions $p(\mathcal{T})$. This has led to variants that model mixtures over multiple initial conditions \cite{vuorio2019multimodalmaml, rusu2019leo, yao2019hierarchically}.

\keypoint{Optimizer} The above parameter-centric methods usually rely on existing optimizers such as SGD with momentum or Adam \cite{kingma2015adam} to refine the initialization when given some new task. Instead, optimizer-centric approaches \cite{andrychowicz2016graddescgraddesc,wichrowska2017learnoptscalegen,ravi2016optimization,li2017learntooptimize} focus on learning the inner optimizer by training a function that takes as input optimization states such as $\theta$ and $\nabla_\theta\mathcal{L}^{task}$ and produces the optimization step for each base learning iteration. The trainable component $\omega$ can span simple hyper-parameters such as a fixed step size \cite{antoniou2018maml,li2017meta} to more sophisticated pre-conditioning matrices \cite{park2019metacurvature,flennerhag2020warpgrad}. Ultimately $\omega$ can be used to define a full gradient-based optimizer through a complex non-linear transformation of the input gradient and other metadata \cite{ravi2016optimization,andrychowicz2016graddescgraddesc,wichrowska2017learnoptscalegen,bello2017neuraloptimizerrl}. The parameters to learn here can be few if the optimizer is applied coordinate-wise across weights \cite{andrychowicz2016graddescgraddesc}. The initialization-centric and optimizer-centric methods can be merged by learning them jointly, namely having the former learn the initial condition for the latter \cite{ravi2016optimization,li2017meta}. Optimizer learning methods have both been applied to for few-shot learning \cite{ravi2016optimization} and to accelerate and improve many-shot learning \cite{andrychowicz2016graddescgraddesc,wichrowska2017learnoptscalegen,bello2017neuraloptimizerrl}.
Finally, one can also meta-learn zeroth-order optimizers \cite{chen2017l2lgdgd} that only require evaluations of $\mathcal{L}^{task}$ rather than optimizer states such as gradients. These have been shown \cite{chen2017l2lgdgd} to be competitive with conventional Bayesian Optimization \cite{shahriari2016bayesianoptimizationsurvey} alternatives.

\keypoint{Feed-Forward Models (FFMs. aka, Black-Box, Amortized)}
Another family of models trains learners $\omega$ that provide a feed-forward mapping directly from the support set to the parameters required to classify test instances, i.e., $\theta=g_\omega(\mathcal{D}^{train})$ -- rather than relying on a gradient-based iterative optimization of $\theta$. These correspond to  black-box model-based learning in the conventional taxonomy (Sec.~\ref{sec:alg}) and span from classic \cite{heskes2000empirical} to recent approaches such as CNAPs \cite{requeima2019cnap} that provide strong performance on challenging cross-domain few-shot benchmarks \cite{triantafillou2019metadataset}.

These methods have connections to Hypernetworks \cite{ha2016hypernetworks, brock2017smash} which generate the weights of another neural network conditioned on some embedding -- and are often used for compression or multi-task learning. Here $\omega$ is the hypernetwork and it synthesises $\theta$ given the source dataset in a feed-forward pass \cite{rusu2019leo,rakelly2019pearl}. Embedding the support set is often achieved by recurrent networks \cite{duan2017one,wang2016l2rl,hochreiter2001learning} convolution \cite{mishra2017simple}, or set embeddings \cite{gordon2018metaprobainference,requeima2019cnap}.
Research here often studies architectures for paramaterizing the classifier by the task-embedding network: (i) Which parameters should be globally shared across all tasks, vs synthesized per task by the hypernetwork (e.g., share the feature extractor and synthesize the classifier \cite{chen2019closerfewshot,qiao2017parametersfromactivations}), and (ii) How to parameterize the hypernetwork so as to limit the number of parameters required in $\omega$ (e.g., via synthesizing only lightweight adapter layers in the feature extractor \cite{requeima2019cnap}, or class-wise classifier weight synthesis  \cite{gordon2018metaprobainference}). 

Some FFMs can also be understood elegantly in terms of amortized inference in probabilistic models \cite{heskes2000empirical,gordon2018metaprobainference}, making predictions for test data $x$ as:
\begin{equation}
q_\omega(y|x,\mathcal{D}^{tr})=\int p(y|x,\theta)q_\omega(\theta|\mathcal{D}^{tr})d\theta \label{eq:amortized}
\end{equation}
where the meta-representation $\omega$ is a network $q_\omega(\cdot)$ that approximates the intractable Bayesian inference for parameters $\theta$ that solve the task with training data $\mathcal{D}^{tr}$, and the integral may be computed exactly \cite{heskes2000empirical}, or approximated by sampling \cite{gordon2018metaprobainference} or point estimate \cite{requeima2019cnap}. The model $\omega$ is then trained to minimise validation loss over a distribution of training tasks \emph{cf.} Eq.~\ref{eq:metaBBM}.

Finally, memory-augmented neural networks, with the ability to remember old data and assimilate new data quickly, typically fall in the FFM category as well \cite{santoro2016metamemory,munkhdalai2017metanets}.

\keypoint{Embedding Functions (Metric Learning)}
Here the meta-optimization process learns an embedding network $\omega$ that transforms raw inputs into a representation suitable for recognition by simple similarity comparison between query and support instances \cite{qiao2017parametersfromactivations,chen2019closerfewshot,vinyals2016matching,snell2017prototypicalnets} (e.g., with cosine similarity or euclidean distance). These methods are classified as metric learning in the conventional taxonomy (Section~\ref{sec:alg}) but can also be seen as a special case of the feed-forward black-box models above. This can easily be seen for methods that produce logits based on the inner product of the embeddings of support and query images $x_s$ and $x_q$, namely $g^T_\omega(x_{q}) g_\omega(x_{s})$ \cite{qiao2017parametersfromactivations,chen2019closerfewshot}. Here the support image generates `weights' to interpret the query example, making it a special case of a FFM where the `hypernetwork' generates a linear classifier for the query set. Vanilla methods in this family have been further enhanced by making the embedding task-conditional \cite{oreshkin2018tadam,antoniou2019learning}, learning a more elaborate comparison metric \cite{sung2018relationnet,garcia2018fewshot}, or combining with gradient-based meta-learning to train other hyper-parameters such as stochastic regularizers \cite{tseng2020crossdomain}.

\keypoint{Losses and Auxiliary Tasks} Analogously to the meta-learning approach to optimizer design, these  aim to learn the inner task-loss $\mathcal{L}_\omega^{task}(\cdot)$ for the base model. Loss-learning approaches typically define a small neural network that inputs quantities relevant to losses (e.g. predictions, features, or model parameters) and outputs a scalar to be treated as a loss by the inner (task) optimizer. This has potential benefits such as leading to a learned loss that is \emph{easier} to optimize (e.g. less local minima) than commonly used ones \cite{houthooft2018epg,sung2017learning,zhou2020online}, leads to faster learning with improved generalization \cite{denevi2018l2lmean,denevi2019online,gonzalez2019baikalloss,bechtle2019metaLoss}, or one whose minima correspond to a model more robust to domain shift \cite{li2019featurecritic}. Loss learning methods have also been used to learn to learn from unlabeled instances \cite{antoniou2019learning,semifewmaml2018}, or to learn $\mathcal{L}_\omega^{task}()$ as a differentiable approximation to a true non-differentiable task loss such as area under precision recall curve \cite{huang2019lossmetric,grabocka2019surrogateloss}.

Loss learning also arises in generalizations of self-supervised
\cite{doersch2017mtlselfsup} or auxiliary task \cite{jaderberg2017unsupauxrl} learning. In these problems unsupervised predictive tasks (such as colourising pixels in vision \cite{doersch2017mtlselfsup}, or simply changing pixels in RL \cite{jaderberg2017unsupauxrl}) are defined and optimized 
with the aim of improving the representation for the main task. In this case the best auxiliary task (loss) to use can be hard to predict in advance, so meta-learning can be used to select among several auxiliary losses according to their impact on improving main task learning. I.e., $\omega$ is a per-auxiliary task weight \cite{xingyu2019auxtask}. More generally, one can meta-learn an auxiliary task generator that annotates examples with auxiliary labels \cite{liu2019self}.

\keypoint{Architectures}
Architecture discovery has always been an important area in neural networks \cite{stanley2019designing,elsken2019nas}, and one that is not amenable to simple exhaustive search. Meta-Learning can be used to automate this very expensive process by learning architectures. Early attempts used evolutionary algorithms to learn the topology of LSTM cells \cite{bayer2009evolvingLSTMcells}, while later approaches leveraged RL to generate descriptions for good CNN architectures \cite{zoph2016neural}. Evolutionary Algorithms \cite{real2018regularized} can learn blocks within architectures modelled as graphs which could mutate by editing their graph. Gradient-based architecture representations have also been visited in the form of DARTS \cite{liu2019darts} where the forward pass during training consists in a softmax across the outputs of all possible layers in a given block, which are weighted by coefficients to be meta learned (i.e. $\omega$). During meta-test, the architecture is discretized by only keeping the layers corresponding to the highest coefficients.
Recent efforts to improve DARTS have focused on more efficient differentiable approximations \cite{xie2019snas}, robustifying the discretization step \cite{Zela2020robustifyingDARTS}, learning easy to adapt initializations \cite{lian2020towards}, or architecture priors \cite{shaw2019meta}. See Section~\ref{subsec:nasApplication} for more details.

\keypoint{Attention Modules} 
have been used as comparators in metric-based meta-learners \cite{hou2019cross}, to prevent catastrophic forgetting in few-shot continual learning \cite{ren2019incremental}  and to summarize the distribution of text classification tasks \cite{bao2019few}.

\keypoint{Modules} Modular meta-learning \cite{alet2018modularmetalearning,alet2019modularmeta} assumes that the task agnostic knowledge $\omega$ defines a set of modules, which are re-composed in a task specific manner defined by $\theta$ in order to solve each encountered task. These strategies can be seen as meta-learning generalizations of the typical structural approaches to knowledge sharing that are well studied in multi-task and transfer learning \cite{yang2017deepmtrl,fernando2017pathnet,meyerson2019modular}, and may ultimately underpin compositional learning \cite{lake2019compositional}.

\keypoint{Hyper-parameters} Here $\omega$ represents hyperparameters of the base learner such as regularization strength \cite{franceschi2018bilevel, micaelli2020nongreedy}, per-parameter regularization \cite{balaji2018metareg}, task-relatedness in multi-task learning \cite{franceschi2017hyperopt}, or sparsity strength in data cleansing \cite{franceschi2017hyperopt}. Hyperparameters such as step size \cite{li2017meta,antoniou2018maml, micaelli2020nongreedy} can be seen as part of the optimizer, leading to an overlap between hyper-parameter and optimizer learning categories.

\keypoint{Data Augmentation} In supervised learning it is common to improve generalization by synthesizing more training data through label-preserving transformations on the existing data. The data augmentation operation is wrapped up in optimization steps of the inner problem (Eq.~\ref{eq:lower}), and is conventionally hand-designed. However, when $\omega$ defines the data augmentation strategy, it can be learned by the outer optimization in Eq.~\ref{eq:upper} in order to maximize validation performance \cite{cubuk2019autoaug}. Since augmentation operations are typically non-differentiable, this requires reinforcement learning \cite{cubuk2019autoaug}, discrete gradient-estimators \cite{li2020dada}, or evolutionary \cite{volpi2019learnaugment} methods. An open question is whether powerful GAN-based data augmentation methods \cite{antoniou2017data} can be used in inner-level learning and optimized in outer-level learning.

\keypoint{Minibatch Selection, Sample Weights, and Curriculum Learning}
When the base algorithm is minibatch-based stochastic gradient descent, a design parameter of the learning strategy is the batch selection process. Various hand-designed methods \cite{zhang2019active} exist to improve on randomly-sampled minibatches. Meta-learning approaches can define $\omega$ as an instance selection probability \cite{loshchilov2016onlinebatch} or neural network that picks instances \cite{fan2018learnteach} for inclusion in a minibatch. Related to mini-batch selection policies are methods that learn \emph{per-sample} loss weights $\omega$ for the training set \cite{shu2019metaweightnet,ren2018l2rw}. This can be used to learn under label-noise by discounting noisy samples \cite{shu2019metaweightnet,ren2018l2rw}, discount outliers \cite{franceschi2017hyperopt}, or correct for class imbalance \cite{shu2019metaweightnet}

More generally, the \emph{curriculum} \cite{elman1993startingsmall} refers to sequences of data or concepts to learn that produce better performance than learning items in a random order. For instance by focusing on instances of the right difficulty while rejecting too hard or too easy (already learned) instances. Instead of defining a curriculum by hand  \cite{bengio2009curriculumlearning}, meta-learning can automate the process and select examples of the right difficulty by defining a teaching policy as the meta-knowledge and training it to optimize the student's progress  \cite{jiang2018mentornet,fan2018learnteach}.

\keypoint{Datasets, Labels and Environments}
Another meta-representation is the support dataset itself. This departs from our initial formalization of meta-learning which considers the source datasets to be fixed (Section~\ref{sec:formalize}, Eqs.~\ref{eq:metaAbstract}-\ref{eq:metatrain}). However, it can be easily understood in the bilevel view of Eqs.~\ref{eq:upper}-\ref{eq:lower}. If the validation set in the upper optimization is real and fixed, and a train set in the lower optimization is paramaterized by $\omega$, the training dataset can be tuned by meta-learning to optimize validation performance.

In dataset distillation \cite{wang2018datasetdistill, lorraine2019optmillionsofhyperparams}, the support images themselves are learned such that a few steps on them allows for good generalization on real query images. This can be used to summarize large datasets into a handful of images, which is useful for replay in continual learning where streaming datasets cannot be stored.

Rather than learning input images $x$ for fixed labels $y$, one can also learn the input labels $y$ for fixed images $x$. This can be used in distilling core sets \cite{bohdal2020flexible} as in dataset distillation; or semi-supervised learning, for example to directly learn the unlabeled set's labels to optimize validation set performance \cite{weihong2019learn2impute, sun2019learning}.

In the case of sim2real learning \cite{openai2018learning} in computer vision or reinforcement learning, one uses an environment simulator to generate data for training. In this case, as detailed in Section~\ref{sec:sim2real}, one can also train the graphics engine \cite{ruiz2019learn2sim} or simulator \cite{vuong2019howtorandomize} so as to optimize the real-data (validation) performance of the downstream model after training on data generated by that environment simulator.

\keypoint{Discussion: Transductive Representations and Methods} Most of the representations $\omega$ discussed above are parameter vectors of functions that process or generate data. However a few of the representations mentioned are transductive in the sense that the $\omega$ literally corresponds to data points \cite{wang2018datasetdistill}, labels \cite{weihong2019learn2impute}, or per-sample weights \cite{ren2018l2rw}. Therefore the number of parameters in $\omega$ to meta-learn scales as the size of the dataset. While the success of these methods is a testament to the capabilities of contemporary meta-learning \cite{lorraine2019optmillionsofhyperparams}, this property may ultimately limit their scalability.

Distinct from a transductive representation are methods that are transductive in the sense that they operate on the query instances as well as support instances \cite{antoniou2019learning,liu2019self}.

\keypoint{Discussion: Interpretable Symbolic Representations} A cross-cutting distinction that can be made across many of the meta-representations discussed above is between uninterpretable (sub-symbolic) and human interpretable (symbolic) representations. Sub-symbolic representations, such as when $\omega$ parameterizes a neural network \cite{andrychowicz2016graddescgraddesc}, are more common and make up the majority of studies cited above. However, meta-learning with symbolic representations is also possible, where $\omega$ represents human readable symbolic functions such as optimization program code \cite{bello2017neuraloptimizerrl}. 
Rather than neural loss functions \cite{li2019featurecritic}, one can train symbolic losses $\omega$ that are defined by an expression analogous to cross-entropy \cite{gonzalez2019baikalloss}. One can also meta-learn new symbolic activations \cite{ramachandran2017swish} that outperform standards such as ReLU. As these meta-representations are non-smooth, the meta-objective is non-differentiable and is harder to optimize (see Section~\ref{sec:metaOpt}). So the upper optimization for $\omega$ typically uses RL \cite{bello2017neuraloptimizerrl} or evolutionary algorithms \cite{gonzalez2019baikalloss}. However, symbolic representations may have an advantage \cite{bello2017neuraloptimizerrl,gonzalez2019baikalloss,ramachandran2017swish} in their ability to generalize across task families. I.e., to span wider distributions $p(\mathcal{T})$ with a single $\omega$ during meta-training, or to have the learned $\omega$ generalize to an out of distribution task during meta-testing (see Section~\ref{sec:challenges}). 

\keypoint{Discussion: Amortization} One way to relate some of the representations discussed is in terms of the degree of learning \emph{amortization} entailed \cite{gordon2018metaprobainference}. That is, how much task-specific optimization is performed during meta-testing vs  how much learning is amortized during meta-training. Training from scratch, or conventional fine-tuning \cite{yosinski2014howtransferable} perform full task-specific optimization at meta-testing, with no amortization. MAML \cite{finn2017maml} provides limited amortization by fitting an initial condition, to enable learning a new task by \emph{few-step} fine-tuning. Pure FFMs 
\cite{snell2017prototypicalnets,vinyals2016matching,requeima2019cnap} are fully amortized, with no task-specific optimization, and thus enable the fastest learning of new tasks. Meanwhile some hybrid approaches \cite{rusu2019leo,antoniou2019learning,triantafillou2019metadataset,lee2020balance} implement semi-amortized learning by drawing on both feed-forward and optimization-based meta-learning in a single framework.

\begin{table*}[t]
\centering
\resizebox{0.8\textwidth}{!}{
\begin{tabular}{ l | c c c }
 \hline
 \textbf{Meta-Representation}
 & \multicolumn{3}{c}{\textbf{Meta-Optimizer}} \\
 & \textbf{Gradient} & \textbf{RL} & \textbf{Evolution} \\
 \hline

 \rowcolor{Gray}
 \textbf{Initial Condition} & \red{\cite{finn2017maml,finn2018metauniversal,li2017meta,lee2019metaopt,sun2018metatransferfsl,rajeswaran2019metaimplicit,lee2019metaopt,bertinetto2018closedformmeta}} & \red{\cite{liu2019tamingmaml,rothfuss2019promp,fakoor2020metaq}} \red{\cite{finn2017maml,alshedivat2018nonstationarymeta,ritter2018episodicmeta}} & \red{\cite{song2019esmaml,fernando2018baldwinmeta}} \\

 \textbf{Optimizer} & \green{\cite{andrychowicz2016graddescgraddesc,wichrowska2017learnoptscalegen}} \red{\cite{ravi2016optimization,li2017meta,vuorio2018meta,park2019metacurvature,flennerhag2020warpgrad,metz2018meta}} & \green{\cite{li2017learntooptimize,bello2017neuraloptimizerrl}} & \\

 \rowcolor{Gray}
 \textbf{Hyperparam} & \purple{\cite{franceschi2018bilevel, franceschi2017hyperopt}}\green{\cite{micaelli2020nongreedy}} & \red{\cite{xu2018metagradient,young2019metaac}} & \red{\cite{fernando2018baldwinmeta}}\purple{\cite{jaderberg2019poprl}}\\

 \textbf{Feed-Forward model} & \red{\cite{mishra2017simple,santoro2016metamemory,perezrua2020once,garnelo2018conditionalneuralprocesses,gordon2018metaprobainference,requeima2019cnap}}\green{\cite{pakman2019ncpAmortizedCluster,lee2019setTransformer,lee2019deepAmortizedClustering}} & \red{\cite{duan2016fastslowrl,wang2016l2rl,rakelly2019pearl}} & \\

\rowcolor{Gray}
 \textbf{Metric} & \red{\cite{vinyals2016matching,snell2017prototypicalnets,sung2018relationnet}} && \\

 \textbf{Loss/Reward} & \blue{\cite{li2019featurecritic,balaji2018metareg}} \purple{\cite{grabocka2019surrogateloss}}\green{\cite{bechtle2019metaLoss}} & \purple{\cite{huang2019lossmetric}}\red{\cite{veeriah2019discovery,zhou2020online}}\green{\cite{bechtle2019metaLoss}} & \green{\cite{gonzalez2019baikalloss}}\red{\cite{houthooft2018epg}}\purple{\cite{jaderberg2019poprl}}\\

\rowcolor{Gray}
 \textbf{Architecture} & \purple{\cite{liu2019darts}}\red{\cite{lian2020towards}} & \purple{\cite{zoph2016neural}} & \purple{\cite{real2018regularized}}\\

 \textbf{Exploration Policy} & & \red{\cite{zheng2018intrinsicrewardpg,xu2018metaexplore,stadie2018metaexplore,alet2020metacuriosity,garcia2019metamdp,gupta2018metaexplore}} & \\

 \rowcolor{Gray}
 \textbf{Dataset/Environment} & \red{\cite{wang2018datasetdistill}}\purple{\cite{weihong2019learn2impute}} & \blue{\cite{ruiz2019learn2sim}} & \blue{\cite{vuong2019howtorandomize}} \\

 \textbf{Instance Weights} & \purple{\cite{shu2019metaweightnet,ren2018l2rw,jiang2018mentornet}} & &\\

 \rowcolor{Gray}
 \textbf{Feature/Metric} & \red{\cite{vinyals2016matching,snell2017prototypicalnets,sung2018relationnet,garcia2018fewshot}} & &\\

 \textbf{Data Augmentation/Noise} & \purple{\cite{li2020dada}}\blue{\cite{tseng2020crossdomain}}\red{\cite{lee2020metadropout}} & \purple{\cite{cubuk2019autoaug}}&\purple{\cite{volpi2019learnaugment}}\\

 \rowcolor{Gray}
 \textbf{Modules} & \red{\cite{alet2018modularmetalearning,alet2019modularmeta}} & &\\
 \textbf{Annotation Policy} & \red{\cite{bachman2017laalicml,konyushkova2017learning}} & \blue{\cite{pang2018metatransferal}} &\\
 \hline

\end{tabular}
}
\caption{Research papers according to our taxonomy. We use color to indicate salient meta-objective or application goal. We focus on the main goal of each paper for simplicity. The color code is: \red{sample efficiency} (red), \green{learning speed} (green), \purple{asymptotic performance} (purple), \blue{cross-domain} (blue).}
\end{table*}

\subsection{Meta-Optimizer}\label{sec:metaOpt}
Given a choice of which facet of the learning strategy to optimize, the next axis of meta-learner design is actual outer (meta) optimization strategy to use for training $\omega$.

\keypoint{Gradient} A large family of methods use gradient descent on the meta parameters $\omega$ \cite{finn2017maml,ravi2016optimization,franceschi2017hyperopt,li2019featurecritic}. This requires computing derivatives $d\mathcal{L}^{meta}/d\omega$ of the outer objective, which are typically connected via the chain rule to the model parameter $\theta$, $d\mathcal{L}^{meta}/d\omega=(d\mathcal{L}^{meta}/d\theta) (d\theta/d\omega)$. These methods are potentially the most efficient as they exploit analytical gradients of $\omega$. However key challenges include: (i) Efficiently differentiating through many steps of inner optimization, for example through careful design of differentiation algorithms \cite{franceschi2018bilevel,maclaurin2015gradienthyper, micaelli2020nongreedy} and implicit differentiation \cite{rajeswaran2019metaimplicit,russell2019fixingimplicit,lorraine2019optmillionsofhyperparams}, and dealing tractably with the required second-order gradients \cite{nichol2018reptilefoml}. (ii) Reducing the inevitable gradient degradation problems whose severity increases with the number of inner loop optimization steps. (iii) Calculating gradients when the base learner, $\omega$, or $\mathcal{L}^{task}$ include discrete or other non-differentiable operations.

\keypoint{Reinforcement Learning} When the base learner includes non-differentiable steps \cite{cubuk2019autoaug}, or the meta-objective $\mathcal{L}^{meta}$ is itself non-differentiable \cite{huang2019lossmetric}, many methods \cite{duan2016fastslowrl} resort to RL to optimize the outer objective Eq.~\ref{eq:upper}. This estimates the gradient $\nabla_\omega\mathcal{L}^{meta}$, typically using the policy gradient theorem. However, alleviating the requirement for differentiability in this way is typically extremely costly. High-variance policy-gradient estimates for $\nabla_\omega\mathcal{L}^{meta}$ mean that many outer-level optimization steps are required to converge, and each of these steps are themselves costly due to wrapping task-model optimization within them.

\keypoint{Evolution} Another approach for optimizing the meta-objective are evolutionary algorithms (EA) \cite{schmidhuber1987evolutionary,salimans2017evolution,stanley2019designing}. Many evolutionary algorithms have strong connections to reinforcement learning algorithms \cite{stulp2013rles}. However, their performance does not depend on the length and reward sparsity of the inner optimization as for RL.

EAs are attractive for several reasons \cite{salimans2017evolution}: (i) They can optimize any base model and meta-objective with no differentiability constraint. (ii) Not relying on backpropagation avoids both gradient degradation issues and the cost of high-order gradient computation of conventional gradient-based methods. (iii) They are highly parallelizable for scalability. (iv) By maintaining a diverse population of solutions, they can avoid local minima that plague gradient-based methods \cite{stanley2019designing}. However, they have a number of disadvantages: (i) The population size required increases rapidly with the number of parameters to learn. (ii) They can be sensitive to the mutation strategy  and may require careful hyperparameter optimization. (iii) Their fitting ability is generally inferior to gradient-based methods, especially for large models such as CNNs.

EAs are relatively more commonly applied in RL applications \cite{houthooft2018epg,song2019esmaml} (where models are typically smaller, and inner optimizations are long and non-differentiable). However they have also been applied to learn learning rules \cite{soltoggio2018born}, optimizers \cite{cao2019learning}, architectures \cite{real2018regularized,stanley2019designing} and data augmentation strategies \cite{volpi2019learnaugment} in supervised learning. They are also particularly important in learning human interpretable symbolic meta-representations \cite{gonzalez2019baikalloss}.

\subsection{Meta-Objective and Episode Design}\label{sec:why} The final component is to define the meta-learning goal through
choice of meta-objective $\mathcal{L}^{meta}$, and associated data flow between inner loop episodes and outer optimizations. Most methods define a meta-objective using a performance metric computed on a validation set, after updating the task model with $\omega$. This is in line with classic validation set approaches to hyperparameter and model selection. However, within this framework, there are several design options:

\keypoint{Many vs Few-Shot Episode Design} According to whether the goal is improving few- or many-shot performance, inner loop learning episodes may be defined with many \cite{franceschi2017hyperopt,wichrowska2017learnoptscalegen,bello2017neuraloptimizerrl} or few- \cite{finn2017maml,ravi2016optimization} examples per-task.

\keypoint{Fast Adaptation vs Asymptotic Performance} When validation loss is computed at the end of the inner learning episode, meta-training encourages better \emph{final} performance of the base task. When it is computed as the sum of the validation loss after each inner optimization step, then meta-training also encourages \emph{faster} learning in the base task \cite{antoniou2018maml,wichrowska2017learnoptscalegen,bello2017neuraloptimizerrl}. Most RL applications also use this latter setting.

\keypoint{Multi vs Single-Task} When the goal is to tune the learner to better solve any task drawn from a given family, then inner loop learning episodes correspond to a randomly drawn task from $p(\mathcal{T})$ \cite{finn2017maml,li2019featurecritic,snell2017prototypicalnets}. When the goal is to tune the learner to simply solve one specific task better, then the inner loop learning episodes all draw data from the same underlying task \cite{xu2018metagradient,franceschi2017hyperopt,andrychowicz2016graddescgraddesc,zheng2018intrinsicrewardpg,veeriah2019discovery,meier2018onlinelearningrate}.

It is worth noting that these two meta-objectives tend to have different assumptions and value propositions. The multi-task objective obviously requires a task family $p(\mathcal{T})$ to work with, which single-task does not. Meanwhile for multi-task, the data and compute cost of meta-training can be amortized by potentially boosting the performance of multiple target tasks during meta-test; but single-task -- without the new tasks for amortization -- needs to improve the final solution or asymptotic performance of the current task, or meta-learn fast enough to be online.

\keypoint{Online vs Offline} While the classic meta-learning pipeline defines the meta-optimization as an outer-loop of the inner base learner \cite{finn2017maml,andrychowicz2016graddescgraddesc}, some studies have attempted to preform meta-optimization \emph{online} within a single base learning episode \cite{veeriah2019discovery,meier2018onlinelearningrate,li2019featurecritic, baydin2017hd}. In this case the base model $\theta$ and learner $\omega$ co-evolve during a single episode. Since there is now no set of source tasks to amortize over, meta-learning needs to be fast compared to base model learning in order to benefit sample or compute efficiency.

\keypoint{Other Episode Design Factors} Other operators can be inserted into the episode generation pipeline to customize meta-learning for particular applications. For example one can simulate domain-shift between training and validation to meta-optimize for good performance under domain-shift \cite{balaji2018metareg,li2019featurecritic,li2020metadomain}; simulate network compression such as quantization \cite{chen2019metaquant} between training and validation to meta-optimize for network compressibility; provide noisy labels during meta-training to optimize for label-noise robustness \cite{li2019learning}, or generate an adversarial validation set to meta-optimize for adversarial defense \cite{goldblum2019adversarially}. These opportunities are explored in more detail in the following section.

\section{Applications}

In this section we briefly review the ways in which meta-learning has been exploited in computer vision, reinforcement learning, architecture search, and so on. 
\subsection{Computer Vision and Graphics}
Computer vision is a major consumer domain of meta-learning techniques, notably due to its impact on few-shot learning, which holds promise to deal with the challenge posed by the long-tail of concepts to recognise in vision.

\subsubsection{Few-Shot Learning Methods}\label{sec:fsl}
Few-shot learning (FSL) is extremely challenging, especially for large neural networks \cite{krizhevsky2012imagenetdeepcnn,he2016resnet}, where data volume is often the dominant factor in performance \cite{sun2017unreasonable}, and training large models with small datasets leads to overfitting or  non-convergence. Meta-learning-based approaches are increasingly able to train powerful CNNs on small datasets in many vision problems. We provide a non-exhaustive representative summary as follows.

\keypoint{Classification} The most common application of meta-learning is few-shot multi-class image recognition, where the inner and outer loss functions are typically the cross entropy over training and validation data respectively \cite{vinyals2016matching, snell2017prototypicalnets, ravi2016optimization, li2017meta, antoniou2018maml, garcia2018fewshot, yin2019meta, rusu2019leo, yoon2019tapnet, antoniou2019learning, yao2019hierarchically, rae2019meta, raghu2019rapid, yao2020automated, flennerhag2020warpgrad}. Optimizer-centric \cite{finn2017maml}, black-box \cite{mishra2017simple,qiao2017parametersfromactivations} and metric learning \cite{vinyals2016matching,sung2018relationnet,garcia2018fewshot} models have all been considered. \cut{Relevant benchmarks are covered in Section~\ref{sec:ClassificationBenchmarks}.}

This line of work has led to a steady improvement in performance compared to early methods \cite{vinyals2016matching,finn2017maml,kosh2015siameseoneshot}. However, performance is still far behind that of fully supervised methods, so there is more work to be done. Current research issues include improving cross-domain generalization \cite{tseng2020crossdomain}, recognition within the joint label space defined by meta-train and meta-test classes \cite{gidaris2018dynamicfewshot}, and incremental addition of new few-shot classes \cite{ren2019incremental,perezrua2020once}.

\keypoint{Object Detection} Building on progress in few-shot classification, few-shot object \emph{detection} \cite{perezrua2020once,kang2019fsldet} has been demonstrated, often using feed-forward hypernetwork-based approaches to embed support set images and synthesize final layer classification weights in the base model.

\keypoint{Landmark Prediction} aims to locate a skeleton of key points within an image, such as such as joints of a  human or robot. This is typically formulated as an image-conditional regression. For example, a MAML-based model was shown to work for human pose estimation \cite{gui2018motionprediction}, modular-meta-learning was successfully applied to robotics \cite{alet2018modularmetalearning}, while a hypernetwork-based model was applied to few-shot clothes fitting for novel fashion items \cite{perezrua2020once}.

\keypoint{Few-Shot Object Segmentation} is important due to the cost of obtaining pixel-wise labeled images. Hypernetwork-based meta-learners have been applied in the one-shot regime \cite{shaban2017oneshotsegmentation}, and performance was later improved by adapting prototypical networks \cite{dong2018fewshotss}. Other models tackle cases where segmentation has low density \cite{ rakelly2018fewshotsegmentation}.

\keypoint{Image and Video Generation} In \cite{gordon2018metaprobainference} an amortized probabilistic meta-learner is used to generate multiple views of an object from just a single image, generative query networks \cite{eslami2018neuralGQN} render scenes from novel views, and talking faces are generated from little data by learning the initialization of an adversarial model for quick adaptation \cite{zakharov2019talkingheads}. In video domain,  \cite{wang2019few} meta-learns a weight generator that synthesizes videos given few example images as cues.


\keypoint{Generative Models and Density Estimation} Density estimators capable of generating images typically require many parameters, and as such overfit in the few-shot regime. Gradient-based meta-learning of PixelCNN generators was shown to enable their few-shot learning \cite{reed2017fewshotdensityestimation}.

\subsubsection{Few-Shot Learning Benchmarks}\label{sec:ClassificationBenchmarks}
Progress in AI and machine learning is often measured, and spurred, by well designed benchmarks \cite{russakovsky2015imagenet}. Conventional ML benchmarks define a task and dataset for which a model should generalize from seen to unseen \emph{instances}. 
In meta-learning, benchmark design is more complex, since we are often dealing with a learner that should generalize from seen to unseen \emph{tasks}. Benchmark design thus needs to define families of tasks from which meta-training and meta-testing tasks can be drawn. Established FSL benchmarks include miniImageNet \cite{vinyals2016matching,ravi2016optimization}, Tiered-ImageNet \cite{ren2018meta}, SlimageNet \cite{antoniou2020defining}, Omniglot \cite{vinyals2016matching} and Meta-Dataset \cite{triantafillou2019metadataset}. 

\keypoint{Dataset Diversity, Bias and Generalization} The standard benchmarks provide   tasks for training and evaluation, but suffer from a lack of diversity (narrow $p(\mathcal{T})$) which makes performance on these benchmarks non-reflective of performance on real-world few shot task. For example, switching between different kinds of animal photos in miniImageNet is not a strong test of generalization. Ideally we would like to span more diverse categories and types of images (satellite, medical, agricultural, underwater, etc); and even be robust to domain-shifts between meta-train and meta-test tasks.

There is work still to be done here as, even in the many-shot setting, fitting a deep model to a very wide distribution of data is itself non-trivial \cite{rebuff2017mdl}, as is generalizing to out-of-sample data \cite{li2019featurecritic,balaji2018metareg}. Similarly, the performance of meta-learners often drops drastically when introducing a domain shift between the source and target task distributions \cite{chen2019closerfewshot}. This motivates the recent Meta-Dataset \cite{triantafillou2019metadataset} and CVPR cross-domain few-shot challenge \cite{guo2019new}. Meta-Dataset aggregates a number of individual recognition benchmarks to provide a wider distribution of tasks $p(\mathcal{T})$ to evaluate the ability to fit a wide task distribution and generalize across domain-shift. Meanwhile, \cite{guo2019new} challenges methods to generalize from the everyday ImageNet images to medical, satellite and agricultural images. Recent work has begun to try and address these issues by meta-training for domain-shift robustness as well as sample efficiency \cite{tseng2020crossdomain}. Generalization issues also arise in applying models to data from under-represented countries \cite{de2019does}.

\subsection{Meta Reinforcement Learning and Robotics}

Reinforcement learning is typically concerned with learning control policies that enable an agent to obtain high reward after performing a sequential action task within an environment. RL typically suffers from extreme sample inefficiency due to sparse rewards, the need for exploration, and the  high-variance \cite{williams1992simple} of optimization algorithms.
However, applications often naturally entail task families which meta-learning can exploit -- for example locomoting-to or reaching-to different positions \cite{gupta2018metaexplore}, navigating within different environments \cite{mishra2017simple}, traversing different terrains \cite{clavera2019learning}, driving different cars \cite{garcia2019metamdp}, competing with different competitor agents \cite{alshedivat2018nonstationarymeta}, and dealing with different handicaps such as failures in individual robot limbs \cite{clavera2019learning}. Thus RL provides a fertile application area in which meta-learning on task distributions has had significant successes in improving sample efficiency over standard RL algorithms. One can intuitively understand the efficacy of these methods. For instance  meta-knowledge of a maze layout is transferable for all tasks that require navigating within the maze.

\subsubsection{Methods}
Several meta-representations that we have already seen have been explored in RL including learning the initial conditions \cite{finn2017maml,fernando2018baldwinmeta}, hyperparameters \cite{jaderberg2019poprl,fernando2018baldwinmeta}, step directions \cite{li2017meta} and step sizes \cite{young2019metaac}, which enables gradient-based learning to train a neural policy with fewer environmental interactions; and training fast convolutional \cite{mishra2017simple} or recurrent \cite{duan2016fastslowrl,wang2016l2rl} black-box models to embed the experience of a given environment to synthesize a policy.
Recent work has developed improved meta-optimization algorithms \cite{song2019esmaml,rothfuss2019promp,liu2019tamingmaml} for these tasks, and provided theoretical guarantees for meta-RL \cite{fallah2020provably}.

\keypoint{Exploration}
A meta-representation rather unique to RL is the exploration policy. RL is complicated by the fact that the data distribution is not fixed, but varies according to the agent's actions. Furthermore, sparse rewards may mean that an agent must take many actions before achieving a reward that can be used to guide learning. As such, how to explore and acquire data for learning is a crucial factor in any RL algorithm. Traditionally exploration is based on sampling random actions \cite{schulman2017ppo}, or hand-crafted heuristics \cite{sigaud2019policy}. Several meta-RL studies have instead explicitly treated exploration strategy or curiosity function as meta-knowledge $\omega$; and modeled their acquisition as a meta-learning problem \cite{schmidhuber1997s,garcia2019metamdp,stadie2018metaexplore,alet2020metacuriosity} -- leading to  sample efficiency improvements by `learning how to explore'.

\keypoint{Optimization}
RL is a difficult optimization problem where the learned policy is usually far from optimal, even on `training set' episodes. This means that, in contrast to meta-SL, meta-RL methods are more commonly deployed to increase asymptotic  performance \cite{jaderberg2019poprl,houthooft2018epg,veeriah2019discovery} as well as sample-efficiency, and can lead to significantly better solutions overall. The meta-objective of many meta-RL frameworks is the net return of the agent over a full episode, and thus both sample efficient and asymptotically performant learning are rewarded.
Optimization difficulty also means that there has been relatively more work on learning losses (or rewards) \cite{veeriah2019discovery,zhou2020online,kirsch2020improving,bechtle2019metaLoss} which an RL agent should optimize instead of -- or in addition to -- the conventional sparse reward objective. Such learned losses may be easier to optimize (denser, smoother) compared to the true target \cite{houthooft2018epg,kirsch2020improving}. This also links to exploration as reward learning and can be considered to instantiate meta-learning of learning intrinsic motivation \cite{zheng2018intrinsicrewardpg}.

\keypoint{Online meta-RL}
A significant fraction of meta-RL studies addressed the single-task setting, where the meta-knowledge such as loss \cite{zhou2020online,veeriah2019discovery}, reward \cite{jaderberg2019poprl,zheng2018intrinsicrewardpg}, hyperparameters \cite{xu2018metagradient,young2019metaac}, or exploration strategy \cite{xu2018metaexplore} are trained online together with the base policy while learning a single task. These methods thus do not require task families and provide a direct improvement to their respective base learners' performance.

\keypoint{On- vs Off-Policy meta-RL} A major dichotomy in conventional RL is between on-policy and off-policy learning such as PPO \cite{schulman2017ppo} vs SAC \cite{haarnoja2018sac}. Off-policy methods are usually significantly more sample efficient. However, off-policy methods have been harder to extend to meta-RL, leading to  more meta-RL methods being built on on-policy RL methods, thus limiting the absolute performance of meta-RL. Early work in off-policy meta-RL methods has led to strong results \cite{fakoor2020metaq,zhou2020online,kirsch2020improving,rakelly2019pearl}. Off-policy learning also improves the efficiency of the meta-train stage \cite{rakelly2019pearl}, which can be expensive in meta-RL. It also provides new opportunities to accelerate meta-testing by replay buffer sample from meta-training  \cite{fakoor2020metaq}.

\keypoint{Other Trends and Challenges} 
\cite{clavera2019learning} is noteworthy in demonstrating successful meta-RL on a real-world physical robot. Knowledge transfer in robotics is often best studied \emph{compositionally} \cite{kroemer2019robotlearnreview}. E.g., walking, navigating and object pick/place may be subroutines for a room cleaning robot. However, developing meta-learners with effective compositional knowledge transfer  is an open question, with modular meta-learning \cite{alet2019modularmeta} being an option. Unsupervised meta-RL variants aim to perform meta-training without manually specified rewards \cite{jabri2019unsupervised}, or adapt at meta-testing to a changed environment but without new rewards \cite{yang2019norml}. Continual adaptation provides an agent with the ability to adapt to a sequence of tasks within one meta-test episode \cite{clavera2019learning,ritter2018episodicmeta,alshedivat2018nonstationarymeta}, similar to continual learning. Finally, meta-learning has also been applied to imitation \cite{duan2017one} and inverse RL \cite{ghasemipour2019smile}.

\subsubsection{Benchmarks}
Meta-learning benchmarks for RL typically define a family to solve in order to train and evaluate an agent that learns how to learn. These can be tasks (reward functions) to achieve, or domains (distinct environments or MDPs).  

\keypoint{Discrete Control RL} An early meta-RL benchmark for vision-actuated control is the arcade learning environment (ALE) \cite{machado2018revisiting}, which defines a set of classic Atari games split into meta-training and meta-testing. The protocol here is to evaluate return after a fixed number of timesteps in the meta-test environment. \cut{One issue with Atari games is their determinism which means that an open-loop policy is potentially sufficient to solve them, leading to efforts to insert stochasticity \cite{machado2018revisiting}. }
A challenge is the great diversity (wide $p(\mathcal{T})$) across games, which makes successful meta-training hard and leads to limited benefit from knowledge transfer \cite{machado2018revisiting}. Another benchmark \cite{nichol2018learnfast} is based on splitting Sonic-hedgehog levels into meta-train/meta-test. The task distribution here is narrower and beneficial meta-learning is relatively easier to achieve. Cobbe \etal \cite{cobbe2019quantifying} proposed two purpose designed video games for benchmarking meta-RL. CoinRun game \cite{cobbe2019quantifying} provides $2^{32}$ procedurally generated levels of varying difficulty and visual appearance. They show that some $10,000$ levels of meta-train experience are required to generalize reliably to new levels. CoinRun is primarily designed to test direct generalization rather than fast adaptation, and can be seen as providing a distribution over MDP environments to test generalization rather than over tasks to test adaptation. To better test fast learning in a wider task distribution, ProcGen \cite{cobbe2019quantifying} provides a set of 16 procedurally generated games including CoinRun.

\keypoint{Continuous Control RL} While common benchmarks such as gym \cite{gym} have greatly benefited RL research, there is less consensus on meta-RL benchmarks, making existing work hard to compare. Most continuous control meta-RL studies have proposed home-brewed benchmarks that are low dimensional parametric variants of particular tasks such as navigating to various locations or velocities \cite{finn2017maml,rakelly2019pearl}, or traversing different terrains \cite{clavera2019learning}. Several multi-MDP benchmarks \cite{packer2018assessing,zhao2019investigating} have recently been proposed but these primarily test generalization across different environmental perturbations rather than different tasks.
The Meta-World benchmark \cite{yu2019meta} provides a suite of 50 continuous control tasks with state-based actuation, varying from simple parametric variants such as lever-pulling and door-opening. This benchmark should enable more comparable evaluation, and investigation of generalization within and across task distributions. The meta-world evaluation \cite{yu2019meta} suggests that existing meta-RL methods struggle to generalize over wide task distributions and meta-train/meta-test shifts. This may be due to our meta-RL models being too weak and/or benchmarks being too small, in terms of number and coverage tasks, for effective learning-to-learn.
Another recent benchmark suitable for meta-RL is PHYRE \cite{bakhtin2019phyre} which provides a set of 50 vision-based physics task templates which can be solved with simple actions but are likely to require model-based reasoning to address efficiently. These \cut{are organised into 2 difficulty tiers, and } also provide within and cross-template generalization tests.

\keypoint{Discussion} One complication of vision-actuated meta-RL is disentangling visual generalization (as in computer vision) with fast learning of control strategies more generally. For example CoinRun \cite{cobbe2019quantifying} evaluation showed large benefit from standard vision techniques such as batch norm suggesting that perception is a major bottleneck.

\cut{A topical issue in meta-RL is that it is difficult to fit wide meta-train task distributions with multi-task or meta-learning models -- before even getting to the performance of meta-testing on novel tasks. This may be due to our RL models being too weak and/or benchmarks being too small in terms of number of tasks. Even Meta-World, ProcGen and PHYRE have dozens rather than hundreds of tasks like vision benchmarks such as tieredImageNet. While these latest benchmarks are improving, the field would still benefit from still larger benchmarks with controllable generalization gaps. It would also be beneficial to have benchmarks with greater difficulty such as requiring memory and abstract reasoning, to provide opportunities for more abstract strategies to be meta-learned and exploited across tasks.}

\subsection{Environment Learning and Sim2Real}\label{sec:sim2real}
In Sim2Real we are interested in training a model in simulation that is able to generalize to the real-world. The classic domain randomization approach simulates a wide distribution over domains/MDPs, with the aim of training a sufficiently robust model to succeed in the real world -- and has succeeded in both vision \cite{tremblay2018training} and RL \cite{openai2018learning}. Nevertheless tuning the simulation distribution remains a challenge. This leads to a meta-learning setup where the inner-level optimization learns a model in simulation, the outer-level optimization $\mathcal{L}^{meta}$ evaluates the model's performance in the real-world, and the meta-representation $\omega$ corresponds to the parameters of the simulation environment. This paradigm has been used in RL \cite{vuong2019howtorandomize} as well as vision \cite{ruiz2019learn2sim,kar2019metasim}. In this case the source tasks used for meta-train tasks are not a pre-provided data distribution, but paramaterized by omega, $\mathscr{D}_{source}(\omega)$. However, challenges remain in terms of costly back-propagation through a long graph of inner task learning steps; as well as minimising the number of real-world $\mathcal{L}^{meta}$ evaluations in the case of Sim2Real.

\subsection{Neural Architecture Search (NAS)} 
\label{subsec:nasApplication}
Architecture search \cite{stanley2019designing, liu2019darts, zoph2016neural, real2018regularized,elsken2019nas} can be seen as a kind of hyperparameter optimization where $\omega$ specifies the architecture of a neural network. The inner optimization trains networks with the specified architecture, and the outer optimization searches for architectures with good validation performance. NAS methods have been analysed \cite{elsken2019nas} according to `search space', `search strategy', and `performance estimation strategy'. These correspond to the hypothesis space for $\omega$, the meta-optimization strategy, and the meta-objective. NAS is particularly challenging because: (i) Fully evaluating the inner loop is expensive since it requires training a many-shot neural network to completion. This leads to approximations such as sub-sampling the train set, early termination of the inner loop, and interleaved descent on both $\omega$ and $\theta$ \cite{liu2019darts} as in online meta-learning. (ii.) The search space is hard to define, and optimize. This is because most search spaces are broad, and the space of architectures is not trivially differentiable. This leads to reliance on cell-level search \cite{zoph2016neural,liu2019darts} constraining the search space, RL \cite{zoph2016neural}, discrete gradient estimators \cite{xie2019snas} and evolution \cite{real2018regularized,stanley2019designing}.

\keypoint{Topical Issues} While NAS itself can be seen as an instance of hyper-parameter or hypothesis-class meta-learning, it can also interact with meta-learning in other forms. Since NAS is costly, a topical issue is whether discovered architectures can generalize to new problems \cite{zoph2018transferrablearchitecture}. Meta-training across multiple datasets may lead to
improved cross-task generalization of architectures \cite{shaw2019meta}. Finally, one can also define NAS meta-objectives to train an architecture suitable for few-shot learning \cite{kim2018autometa,elsken2019metalearning}. Similarly to fast-adapting initial condition meta-learning approaches such as MAML \cite{finn2017maml}, one can train good initial architectures \cite{lian2020towards} or architecture priors \cite{shaw2019meta} that are easy to adapt towards specific tasks.

\keypoint{Benchmarks} NAS is often evaluated on  CIFAR-10, but it is costly to perform and results are hard to reproduce due to confounding factors such as tuning of hyperparameters \cite{li2019random}. To support reproducible and accessible research, the NASbenches \cite{ying2019nasbench} provide pre-computed performance measures for a large number of network architectures.

\subsection{Bayesian Meta-learning}
Bayesian meta-learning approaches formalize meta-learning via Bayesian hierarchical modelling, and use Bayesian inference for learning rather than direct optimization of parameters. In the meta-learning context, Bayesian learning is typically intractable, and so approximations such as stochastic variational inference or sampling are used.

Bayesian meta-learning importantly provides uncertainty measures for the $\omega$ parameters, and hence measures of prediction uncertainty which can be important for safety critical applications, exploration in RL, and active learning.

A number of authors have explored Bayesian approaches to meta-learning complex neural network models with competitive results.  For example, extending variational autoencoders to model task variables explicitly \cite{edwards2016neuralstat}. Neural Processes \cite{garnelo2018conditionalneuralprocesses} define a feed-forward Bayesian meta-learner inspired by Gaussian Processes but implemented with neural networks. 
Deep kernel learning is also an active research area that has been adapted to the meta-learning setting \cite{tossou2019adaptivedeepkernellearning}, and is often coupled with Gaussian Processes \cite{patacchiola2019deepkernelgp}. In \cite{grant2018bayesmaml} gradient based meta-learning is recast into a hierarchical empirical Bayes inference problem (i.e. prior learning), which models uncertainty in task-specific parameters $\theta$. Bayesian MAML \cite{kim2018bayesianmaml} improves on this model by using a Bayesian ensemble approach that allows non-Gaussian posteriors over $\theta$, and later work removes the need for costly ensembles \cite{gordon2018metaprobainference, ravi2019amortizedbayesianmeta}. In Probabilistic MAML \cite{finn2018probabilistic}, it is the uncertainty in the metaknowledge $\omega$ that is modelled, while a MAP estimate is used for $\theta$. Increasingly, these Bayesian methods are shown to tackle ambiguous tasks, active learning and RL problems.

Separate from the above, meta-learning has also been proposed to aid the Bayesian inference process itself, as in \cite{Wang2020Bayesian} where the authors adapt a Bayesian sampler to provide efficient adaptive sampling methods.

\cut{\keypoint{Benchmarks}
In Bayesian meta-learning, the point is usually to model the uncertainty in the predictions of our meta-learner, and so performance on standard few-shot classification benchmarks doesn't necessarily capture what we care about. For this reason different tasks have been developed in the literature. Bayesian MAML \cite{kim2018bayesianmaml} extends the sinusoidal regression task of MAML \cite{finn2017maml} to make it more challenging. Probabilistic MAML \cite{finn2018probabilistic} provides a suite of 1D toy examples capable of showing model uncertainty and how this uncertainty can be used in an active learning scenario. It also creates a binary classification task from celebA \cite{liu2015faceattributes}, where the positive class is determined by the presence of two facial attributes, but training images show three attributes, thus introducing ambiguity in which two attributes should be classified on. It is observed that sampling $\omega$ can correctly reflect this ambiguity. Active learning toy experiments are also shown in \cite{kim2018bayesianmaml} as well as reinforcement learning applications, and ambiguous one-shot image generation tasks are used in \cite{gordon2018metaprobainference}. Finally, some researchers propose to look at the accuracy v.s. confidence of the meta-learners (i.e. their calibration) \cite{ravi2019amortizedbayesianmeta}.}

\subsection{Unsupervised Meta-Learning} 
There are several distinct ways in which unsupervised learning can interact with meta-learning, depending on whether unsupervised learning in performed in the inner loop or outer loop, and during meta-train vs meta-test.

\keypoint{Unsupervised Learning of a Supervised Learner} The aim here is to learn a supervised learning algorithm (e.g., via MAML \cite{finn2017maml} style initial condition for supervised fine-tuning), but do so without the requirement of a large set of source tasks for meta-training \cite{hsu2018unsupervised,unsupervised2019khoda,antoniou2019assume}. To this end, synthetic source tasks are constructed without supervision via clustering or class-preserving data augmentation, and used to define the meta-objective for meta-training.

\keypoint{Supervised Learning of an Unsupervised Learner} This family of methods aims to meta-train an unsupervised learner. For example, by training the unsupervised algorithm such that it works well for downstream supervised learning tasks. One can train unsupervised learning rules \cite{metz2018meta} or losses \cite{antoniou2019learning,semifewmaml2018} such that downstream supervised learning performance is optimized -- after re-using the unsupervised representation for a supervised task \cite{metz2018meta}, or adapting based on unlabeled data \cite{antoniou2019learning,semifewmaml2018}. Alternatively, when unsupervised tasks such as clustering exist in a family, rather than in isolation, then learning-to-learn of `how-to-cluster' on several source tasks can provide better performance on new clustering tasks in the family \cite{jiang2019metalcluster,garg2018supervising,lee2019setTransformer,lee2019deepAmortizedClustering,pakman2019ncpAmortizedCluster}. The methods in this group that make use of feed-forward models are often known as \emph{amortized clustering} \cite{lee2019setTransformer,lee2019deepAmortizedClustering}, because they amortize the typically iterative computation of clustering algorithms into the cost of training a single inference model, which subsequently performs clustering using a single feed-froward pass. Overall, these methods help to deal with the ill-definedness of the unsupervised learning problem by transforming it into a problem with a clear supervised (meta) objective.

\subsection{Continual, Online and Adaptive Learning}
\keypoint{Continual Learning} refers to the human-like capability of learning tasks presented in sequence. Ideally this is done while exploiting forward transfer so new tasks are learned better given past experience, without forgetting previously learned tasks, and without needing to store past data \cite{chen2018lifelong}. Deep Neural Networks struggle to meet these criteria, especially as they tend to forget information seen in earlier tasks -- a phenomenon known as \emph{catastrophic forgetting}. Meta-learning can include the requirements of continual learning into a meta-objective, for example by defining a sequence of learning episodes in which the support set contains one new task, but the query set contains examples drawn from all tasks seen until now \cite{vuorio2018meta,flennerhag2020warpgrad}. Various meta-representations can be learned to improve continual learning performance, such as weight priors \cite{ren2019incremental}, gradient descent preconditioning matrices \cite{flennerhag2020warpgrad}, or RNN learned optimizers \cite{vuorio2018meta}, or feature representations \cite{javed2019meta}. A related idea is meta-training representations to support local editing updates \cite{sinitsin2020editable} for improvement without interference. 

\keypoint{Online and Adaptive Learning} also consider tasks arriving in a stream, but are concerned with the ability to effectively adapt to the current task in the stream, more than remembering the old tasks. To this end an online extension of MAML was proposed \cite{finn2019online} to perform MAML-style meta-training online during a task sequence. Meanwhile others \cite{ritter2018episodicmeta,alshedivat2018nonstationarymeta,clavera2019learning} consider the setting where meta-training is performed in advance on source tasks, before meta-testing adaptation capabilities on a sequence of target tasks.

\keypoint{Benchmarks}
A number of benchmarks for continual learning work quite well with standard deep learning methods. However, most cannot readily work with meta-learning approaches as their their sample generation routines do not provide a large number of explicit learning sets and an explicit evaluation sets. Some early steps were made towards defining meta-learning ready continual benchmarks in \cite{finn2019online,vuorio2018meta,javed2019meta}, mainly composed of Omniglot and perturbed versions of MNIST. However, most of those were simply tasks built to demonstrate a method. More explicit benchmark work can be found in \cite{antoniou2020defining}, which is built for meta and non meta-learning approaches alike.

\subsection{Domain Adaptation and Domain Generalization}
\label{subsec:domainshift}
Domain-shift refers to the
statistics of data encountered in deployment being different from those used in training. Numerous domain adaptation and generalization algorithms have been studied to address this issue in supervised, unsupervised, and semi-supervised settings \cite{csurka2017domainadaptationbook}.

\keypoint{Domain Generalization}
Domain \emph{generalization} aims to train models with increased robustness to train-test domain shift  \cite{muandet2013domain}, often by exploiting a distribution over training domains. Using a validation domain that is shifted with respect to the training domain \cite{li2017learntogeneralize}, different kinds of meta-knowledge such as regularizers \cite{balaji2018metareg}, losses \cite{li2019featurecritic}, and noise augmentation \cite{tseng2020crossdomain} can be (meta) learned to maximize the robustness of the learned model to train-test domain-shift.

\keypoint{Domain Adaptation}
To improve on conventional domain \emph{adaptation}\cite{csurka2017domainadaptationbook}, meta-learning can be used to define a meta-objective that optimizes the performance of a base unsupervised DA algorithm \cite{li2020metadomain}.

\keypoint{Benchmarks} Popular benchmarks for DA and DG consider image recognition across multiple domains such as photo/sketch/cartoon. \cut{Datasets with multiple domains are often used in order to provide a domain distribution for meta-learning.} PACS \cite{li2017dg} provides a good starter benchmark, with Visual Decathlon \cite{li2019featurecritic,rebuff2017mdl} and Meta-Dataset \cite{triantafillou2019metadataset} providing larger scale alternatives.

\subsection{Hyper-parameter Optimization} Meta-learning address hyperparameter optimization when considering $\omega$ to specify hyperparameters, such as regularization strength or learning rate. There are two main settings: we can learn hyperparameters that improve training over a distribution of tasks, just a single task. The former case is usually relevant in few-shot applications, especially in optimization based methods. For instance, MAML can be improved by learning a learning rate per layer per step \cite{antoniou2018maml}. The case where we wish to learn hyperparameters for a single task is usually more relevant for many-shot applications \cite{lorraine2019optmillionsofhyperparams,micaelli2020nongreedy}, where some validation data can be extracted from the training dataset, as discussed in Section~\ref{sec:formalize}. End-to-end gradient-based meta-learning has already demonstrated promising scalability to millions of parameters (as demonstrated by MAML \cite{finn2017maml} and Dataset Distillation \cite{wang2018datasetdistill, lorraine2019optmillionsofhyperparams}, for example) in contrast to the classic approaches
(such cross-validation by grid or random \cite{bergstra2012randomsearchhyper} search, or Bayesian Optimization \cite{shahriari2016bayesianoptimizationsurvey}) which are typically only successful with dozens of hyper-parameters.

\subsection{Novel and Biologically Plausible Learners}
Most meta-learning work that uses explicit (non feed-forward/black-box) optimization for the base model is based on gradient descent by backpropagation. Meta-learning can define the function class of $\omega$ so as to lead to the discovery of novel learning rules that are unsupervised \cite{metz2018meta} or biologically plausible \cite{bengio1990learning,miconi2018differentiable,miconi2019backpropamine}, making use of ideas less commonly used in contemporary deep learning such as Hebbian updates \cite{miconi2018differentiable} and neuromodulation \cite{miconi2019backpropamine}.

\subsection{Language and Speech}
\keypoint{Language Modelling} Few-shot language modelling increasingly showcases the versatility of meta-learners. Early matching networks showed impressive performances on one-shot tasks such as filling in missing words \cite{vinyals2016matching}. Many more tasks have since been tackled, including text classification \cite{bao2019few}, neural program induction \cite{delvin2017neuralprogram} and synthesis \cite{si2018learning}, English to SQL program synthesis \cite{huang2018naturallanguagequery}, text-based relationship graph extractor \cite{nips2019_9182}, machine translation \cite{gu2018fewshottranslation}, and quickly adapting to new personas in dialogue  \cite{lin2019adaptingnewperson}.

\keypoint{Speech Recognition} Deep learning is now the dominant paradigm for state of the art automatic speech recognition (ASR). Meta-learning is beginning to be applied to address the many few-shot adaptation problems that arise within ASR including learning how to train for low-resource languages \cite{hsu2019metaasr}, cross-accent adaptation \cite{winata2020learning} and optimizing models for individual speakers \cite{klejch2018learning}.

\subsection{Meta-learning for Social Good}
Meta-learning lands itself to various challenging tasks that arise in applications of AI for social good such as medical image classification and drug discovery, where data is often scarce. Progress in the medical domain is especially relevant given the global shortage of pathologists \cite{metter2019pathologistshortage}. In \cite{altae2016oneshotdrugdiscovery} an LSTM is combined with a graph neural network to predict the behaviour of a molecule (e.g. its toxicity) in the one-shot data regime. In \cite{maicas2018maml4radiology} MAML is adapted to weakly-supervised breast cancer detection tasks, and the order of tasks are selected according to a curriculum. MAML is also combined with denoising autoencoders to do medical visual question answering \cite{nguyen2019medicalvqa}, while learning to weigh support samples \cite{ren2018meta} is adapted to pixel wise weighting for skin lesion segmentation tasks that have noisy labels \cite{mirikharaji2019noisyskinlesion}.

\subsection{Abstract Reasoning}
A long- term goal in deep learning is to go beyond simple perception tasks and tackle more abstract reasoning problems such as IQ tests in the form of Raven's Progressive Matrices (RPMs) \cite{barrett2018abstractreasoningiqtest}. Solving RPMs can be seen as asking for few-shot generalization from the context panels to the answer panels. Recent meta-learning approaches to abstract reasoning with RPMs achieved significant improvement via meta-learning a teacher that defines the data generating distribution for the panels \cite{zheng2019abstractreason}. The teacher is trained jointly with the student, and rewarded by the student's progress.

\subsection{Systems}
\keypoint{Network Compression}
Contemporary CNNs require large amounts of memory that may be prohibitive on embedded devices. Thus network compression in various forms such as quantization and pruning are topical research areas \cite{dai2018compressing}. Meta-learning is beginning to be applied to this objective as well, such as training gradient generator meta-networks that allow quantized networks to be trained \cite{chen2019metaquant}, and weight generator meta-networks that allow quantized networks to be trained with gradient \cite{liu2019metapruning}.

\keypoint{Communications} Deep learning is rapidly impacting communications systems. For example by learning coding systems that exceed the best hand designed codes for realistic channels \cite{oshea2017drlphysical}. Few-shot meta-learning can be used to provide rapid  adaptation of codes to changing channel characteristics \cite{jiang2019mind}.

\keypoint{Active Learning (AL)} methods
wrap supervised learning, and define a policy for selective data annotation -- typically in the setting where annotation can be obtained sequentially. The goal of AL is to find the optimal subset of data to annotate so as to maximize performance of downstream supervised learning with the fewest annotations. AL is a well studied problem with numerous hand designed algorithms \cite{settles2012activelearning}. Meta-learning can map active learning algorithm design into a learning task by: (i) defining the inner-level optimization as conventional supervised learning on the annotated dataset so far, (ii) defining $\omega$ to be a query policy that selects the best unlabeled datapoints to annotate, (iii), defining the meta-objective as validation performance after iterative learning and annotation according to the query policy, (iv) performing outer-level optimization to train the optimal annotation query policy \cite{pang2018metatransferal,bachman2017laalicml,konyushkova2017learning}. However, if labels are used to train AL algorithms, they need to generalize across tasks to amortize their training cost \cite{pang2018metatransferal}.

\keypoint{Learning with Label Noise} commonly arises when large datasets are collected by web scraping or crowd-sourcing. While there are many algorithms hand-designed for this situation, recent meta-learning methods have addressed label noise. For example by transductively learning sample-wise weighs to down-weight noisy samples \cite{shu2019metaweightnet}, or learning an initial condition robust to noisy label training \cite{li2019learning}.

\keypoint{Adversarial Attacks and Defenses}
Deep Neural Networks can be fooled into misclassifying a data point that should be easily recognizable, by adding a carefully crafted human-invisible perturbation to the data \cite{goodfellow2015adversarialexamples}. Numerous attack and defense methods have been published in recent years, with defense strategies usually consisting in carefully hand-designed architectures or training algorithms. Analogous to the case in domain-shift, one can train the learning algorithm for robustness by defining a meta-loss in terms of performance under adversarial attack \cite{goldblum2019adversarially,yin2018advmeta}.  

\keypoint{Recommendation Systems}  are a mature consumer of machine learning in the commerce space. However, bootstrapping recommendations for new users with little historical interaction data, or new items for recommendation remains a challenge known as the \emph{cold-start} problem. Meta-learning has applied black-box models to item cold-start \cite{vartak2017coldItem} and gradient-based methods to user cold-start \cite{bharadhwaj2019coldUser}.

\section{Challenges and Open Questions}\label{sec:challenges}

\keypoint{Diverse and multi-modal task distributions} 
{The difficulty of fitting a meta-learner to a distribution of tasks $p(\mathcal{T})$ can depend on its \emph{width}. Many big successes of meta-learning have been within narrow task families, while learning on diverse task distributions can challenge existing methods \cite{yu2019meta,triantafillou2019metadataset,rebuff2017mdl}. This may be partly due to conflicting gradients between tasks \cite{yu2020gradientsurgery}.}

{Many meta-learning frameworks \cite{finn2017maml} implicitly assume that the distribution over tasks $p(\mathcal{T})$ is \emph{uni-modal}, and a single learning strategy $\omega$ provides a good solution for them all. However task distributions are often multi-modal; such as medical vs satellite vs everyday images in computer vision, or putting pegs in holes vs opening doors \cite{yu2019meta} in robotics. Different tasks within the distribution may require different learning strategies, which is hard to achieve with today's methods. In vanilla multi-task learning, this phenomenon is relatively well studied with, e.g., methods that group tasks into clusters \cite{kang2011whosharemtl} or subspaces \cite{yang2015mdmtl}. However this is only just beginning to be explored in meta-learning \cite{allen2019infprototypes}.}

\keypoint{Meta-generalization} {Meta-learning poses a new generalization challenge across tasks analogous to the challenge of generalizing across instances in conventional machine learning. There are two sub-challenges: (i) The first is generalizing from meta-train to novel meta-test tasks drawn from $p(\mathcal{T})$. This is exacerbated because the number of \emph{tasks} available for meta-training is typically low (much less than the number of \emph{instances} available in conventional supervised learning), making it difficult to generalize. One failure mode for generalization in few-shot learning has been well studied under the guise of \emph{memorisation} \cite{yin2019meta}, which occurs when each meta-training task can be solved directly without performing any task-specific adaptation based on the support set. In this case models fail to generalize in meta-testing, and specific regularizers \cite{yin2019meta} have been proposed to prevent this kind of meta-overfitting. 
(ii) The second challenge is generalizing to meta-test tasks drawn from a different distribution than the training tasks. This is inevitable in many potential practical applications of meta-learning, for example generalizing few-shot visual learning from everyday training images of ImageNet to specialist domains such as medical images \cite{guo2019new}. From the perspective of a learner, this is a meta-level generalization of the domain-shift problem, as observed in supervised learning. Addressing these issues through meta-generalizations of regularization, transfer learning, domain adaptation, and domain generalization are emerging directions \cite{tseng2020crossdomain}. Furthermore, we have yet to understand which kinds of meta-representations tend to generalize better under certain types of domain shifts. }

\cut{Another interesting direction could be investigating how introducing yet another level of learning abstractions could affect generalization performance, that is, \emph{meta-meta-learning}. By learning how to do meta-learning, perhaps we can find meta-optimizers that can generalize very strongly across a large variety of types and intensities of domain and even modality shifts. Of course computational costs would grow accordingly}

\keypoint{Task families} Many existing meta-learning frameworks, especially for few-shot learning, require task families for meta-training. While this indeed reflects lifelong human learning, in some applications data for such task families may not be available. Unsupervised meta-learning \cite{hsu2018unsupervised,unsupervised2019khoda,antoniou2019assume} and single-task meta-learning methods \cite{meier2018onlinelearningrate,xu2018metagradient,li2019featurecritic,zheng2018intrinsicrewardpg,veeriah2019discovery}, could help to alleviate this requirement; as can improvements in meta-generalization discussed above.

\keypoint{Computation Cost \& Many-shot}
A naive implementation of bilevel optimization as shown in Section~\ref{sec:formalize} is expensive in both time (because each outer step requires several inner steps) and memory (because reverse-mode differentiation requires storing the intermediate inner states). For this reason, much of meta-learning has focused on the few-shot regime \cite{finn2017maml}. However, there is an increasing focus on methods which seek to extend optimization-based meta-learning to the many-shot regime. Popular solutions include implicit differentiation of $\omega$ \cite{pedregosa2016approxgrads, rajeswaran2019metaimplicit, lorraine2019optmillionsofhyperparams}, forward-mode differentiation of $\omega$ \cite{Williams1989algorithmrecurrentneuralnetwork, franceschi2017hyperopt, micaelli2020nongreedy}, gradient preconditioning \cite{flennerhag2020warpgrad}, solving for a greedy version of $\omega$ online by alternating inner and outer steps \cite{baydin2017hd, liu2019darts, li2019featurecritic}, truncation \cite{shaban2018truncated}, shortcuts \cite{fu2016drmad} or inversion \cite{maclaurin2015gradienthyper} of the inner optimization. Many-step meta-learning can also be achieved by learning an initialization that minimizes the gradient descent trajectory length over task manifolds \cite{flennerhag2019transferring}. Finally, another family of approaches accelerate meta-training via closed-form solvers in the inner loop \cite{bertinetto2018closedformmeta,lee2019metaopt}.

Implicit gradients scale to large dimensions of $\omega$ but only provide approximate gradients for it, and require the inner task loss to be a function of $\omega$. Forward-mode differentiation is exact and doesn't have such constraints, but scales poorly with the dimension of $\omega$. Online methods are cheap but suffer from a short-horizon bias \cite{wu2018shortHorizonBiasL2L}. Gradient degradation is also a challenge in the many-shot regime, and solutions include warp layers \cite{flennerhag2020warpgrad} or gradient averaging \cite{micaelli2020nongreedy}. 

In terms of the cost of solving new tasks at the meta-test stage, FFMs have a significant advantage over optimization-based meta-learners, which makes them appealing for applications involving deployment of learning algorithms 
on mobile devices such as smartphones \cite{ignatov2019aismartphone}, for example to achieve personalisation. This is especially so because the embedded device versions of contemporary deep learning software frameworks typically lack support for backpropagation-based training, which FFMs do not require.

\section{Conclusion}

The field of meta-learning has seen a rapid growth in interest. This has come with some level of confusion, with regards to how it relates to neighbouring fields, what it can be applied to, and how it can be benchmarked. In this survey we have sought to clarify these issues by thoroughly surveying the area both from a methodological point of view -- which we broke down into a taxonomy of meta-representation, meta-optimizer and meta-objective; and from an application point of view. We hope that this survey will help newcomers and practitioners to orient themselves to develop and exploit in this growing field, as well as highlight opportunities for future research.


%



\ifCLASSOPTIONcompsoc
 \section*{Acknowledgments}
\else
 \section*{Acknowledgment}
\fi

T. Hospedales was supported by the Engineering and Physical Sciences Research Council of the UK (EPSRC) Grant number EP/S000631/1 and the UK MOD University Defence Research Collaboration (UDRC) in Signal Processing, and EPSRC Grant EP/R026173/1.

\ifCLASSOPTIONcaptionsoff
 \newpage
\fi



%

\bibliographystyle{IEEEtran}
\bibliography{bibliography}

\newcommand{\picsize}{0.7in}

\begin{IEEEbiography}[{\includegraphics[width=\picsize,height=\picsize,clip,keepaspectratio]{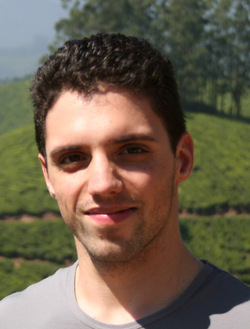}}]{Timothy Hospedales} is a Professor at the University of Edinburgh, and Principal Researcher at Samsung AI Research. His research interest is in data efficient and robust learning-to-learn with diverse applications in vision, language, reinforcement learning, and beyond.
\end{IEEEbiography}

\begin{IEEEbiography}[{\includegraphics[width=\picsize,height=\picsize,clip,keepaspectratio]{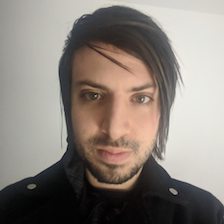}}]{Antreas Antoniou} 
is a PhD student at the University of Edinburgh, supervised by Amos Storkey. His research contributions in meta-learning and few-shot learning are commonly seen as key benchmarks in the field. His main interests lie around meta-learning better learning priors such as losses, initializations and neural network layers, to improve few-shot and life-long learning. 
\end{IEEEbiography}

\begin{IEEEbiography}[{\includegraphics[width=\picsize,height=\picsize,clip,keepaspectratio]{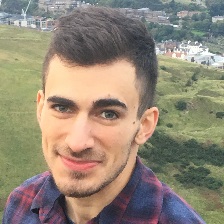}}]{Paul Micaelli} 
is a PhD student at the University of Edinburgh, supervised by Amos Storkey and Timothy Hospedales. His research focuses on zero-shot knowledge distillation and on meta-learning over long horizons for many-shot problems.
\end{IEEEbiography}

\begin{IEEEbiography}[{\includegraphics[width=\picsize,height=\picsize,clip,keepaspectratio]{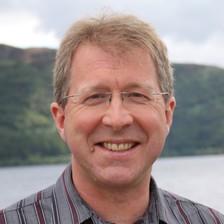}}]{Amos Storkey} 
is Professor of Machine Learning and AI in the School of Informatics, University of Edinburgh. He leads a research team focused on deep neural networks, Bayesian and probabilistic models, efficient inference and meta-learning.
\end{IEEEbiography}

\end{document}